\documentclass{article}

    \PassOptionsToPackage{numbers, compress}{natbib}

\usepackage[main, final]{neurips_2026}

\usepackage[utf8]{inputenc} 
\usepackage[T1]{fontenc}    
\usepackage{hyperref}       
\usepackage{url}            
\usepackage{booktabs}       
\usepackage{amsfonts}       
\usepackage{nicefrac}       
\usepackage{microtype}      
\usepackage[table]{xcolor}  
\usepackage{amsmath}
\usepackage{graphicx}
\usepackage{wrapfig}
\usepackage{algorithm}
\usepackage{algorithmic}
 \usepackage{multirow}
\usepackage{setspace}
\usepackage{tikz}
\usetikzlibrary{positioning,arrows.meta}
\usepackage{minitoc}

\makeatletter
\def\bstctlcite{\@ifnextchar[{\@bstctlcite}{\@bstctlcite[@auxout]}}
\def\@bstctlcite[#1]#2{\@bsphack
  \@for\@citeb:=#2\do{%
    \edef\@citeb{\expandafter\@firstofone\@citeb}%
    \if@filesw\immediate\write\csname #1\endcsname{\string\citation{\@citeb}}\fi}%
  \@esphack}
\makeatother

\newtheorem{theorem}{Theorem}[section]
\newtheorem{proposition}[theorem]{Proposition}

\newtheorem{remark}[theorem]{Remark}

\definecolor{lightgray}{gray}{0.9}  

\title{DiffPTS: Rethinking Diffusion ELBO for Probabilistic Time Series Forecasting}

\author{%
  Weiwei Ye$^{1}$\quad Dongyuan Li$^{1}$\quad Hangchen Liu$^{1}$ \\
  \textbf{Haotong Jiang}$^{2}$\quad \textbf{Yoshihide Sekimoto}$^{1}$\quad \textbf{Renhe Jiang}$^{1,*}$ \\
  $^{1}$The University of Tokyo\qquad $^{2}$Harbin Institute of Technology, Shenzhen\quad $^{*}$Corresponding author
}

\begin{document}
\bstctlcite{IEEEexample:BSTcontrol}

\maketitle

\begin{abstract}
Probabilistic time series forecasting requires modeling and predicting complex and time-varying distributions. Recently, Denoising Diffusion Probabilistic Model~(DDPM)-based approaches have shown promise by equipping the diffusion process with pretrained mean and variance estimators to accommodate distributional shift. However, these methods typically follow the standard DDPM framework and consider only partial components of the evidence lower bound (ELBO), treating the training of estimators as designed regression tasks separate from the variational inference framework.  To address this, we rethink the ELBO under the Location-Scale Noise Model (LSNM) and find that it naturally induces a Gaussian negative log likelihood objective for the estimators and inherently defines a joint training objective that unifies recent diffusion paradigms for probabilistic forecasting. Building on this principled ELBO reformulation, we propose DiffPTS, a general framework that enables end-to-end optimization of all components within the ELBO. Across multiple benchmarks, DiffPTS consistently outperforms recent models, achieving state-of-the-art performance with an average CRPS/MSE reduction of over 14.53\%/16.55\% compared to existing diffusion-based methods. The code is available at \url{https://github.com/wwy155/DiffPTS}.
\end{abstract}

\section{Introduction}
Probabilistic time series forecasting~(PTSF) has become increasingly crucial in various real-world applications ranging from financial risk management~\cite{chen2012bayesian,eggen2025financial, duan2022factorvae} to climate prediction~\cite{eden2015global,palmer2012towards}. In particular, PTSF aims to effectively model the conditional distribution $p(\mathbf{Y}|\mathbf{X})$ of future outcomes $\mathbf{Y}$ given historical observations $\mathbf{X}$~\cite{lim2021time}. Given the dynamic nature of time series data, the fundamental challenge lies in accurately capturing complex and time-varying distributions~\cite{kim2021reversible, ye2025non, zeng2023transformers}.


In recent years, Denoising Diffusion Probabilistic Models (DDPMs)~\cite{ho2020denoising, sohl2015deep} have emerged as a powerful framework for modeling complex distributions, inspiring their growing adoption in probabilistic forecasting scenarios~\cite{tashiro2021csdi, rasul2021autoregressive, shen2023non}. At their core, standard DDPMs optimize the Evidence Lower Bound (ELBO) to learn the data distribution by introducing a denoising network $\epsilon_\theta$ that learns to predict the noise added during the forward diffusion process, enabling tractable variational inference from a fixed Gaussian endpoint $p(\mathbf{Y}_T) =\mathcal{N}(\mathbf{0}, \mathbf{I})$. However, conventional DDPMs often fail in accurate probabilistic forecasting due to distributional shifts~\cite{ye2025non, lee2021priorgrad, li2025diffusion}. To address this limitation, recent approaches have proposed introducing adaptive mean and variance estimators~($f_\phi(\mathbf{X})$, $g_\psi(\mathbf{X}))$  into the diffusion process, either by designing a new residual-based denoising target $(\mathbf{Y} - f_\phi(\mathbf{X}))/g_\psi(\mathbf{X})$  or by modifying the reverse diffusion endpoint to follow $\mathcal{N}(f_\phi(\mathbf{X}), g_\psi(\mathbf{X}))$ to accommodate distributional changes~\cite{ye2025non, lai2025rdit}, which have shown promising results~\cite{lin2024diffusion, su2025diffusion}.

Typically, these advancements follow the standard diffusion training paradigm, and the training of $f_\phi$ and $g_\psi$ is conducted in separate pretraining stages or as auxiliary regression tasks by designing various targets for $f_\phi$ and $g_\psi$ (e.g., NsDiff~\cite{ye2025non} uses a future sliding variance target for $g_\psi$). These diverse design choices for training $f_\phi$ and $g_\psi$ raise a critical question: \textbf{How should $f_\phi$ and $g_\psi$ be trained to best align with the DDPM within the variational inference framework?} Motivated by this question, we rethink the ELBO and surprisingly find that when introducing adaptive estimators into the DDPM, crucial ELBO terms that were previously treated as constants become non-negligible and require explicit joint optimization. Specifically, the reconstruction non-denoising terms and prior matching terms must be strictly optimized to achieve a tight variational bound.

 \begin{wrapfigure}{r}{0.5\linewidth}
  \centering
    \includegraphics[width=1\linewidth]{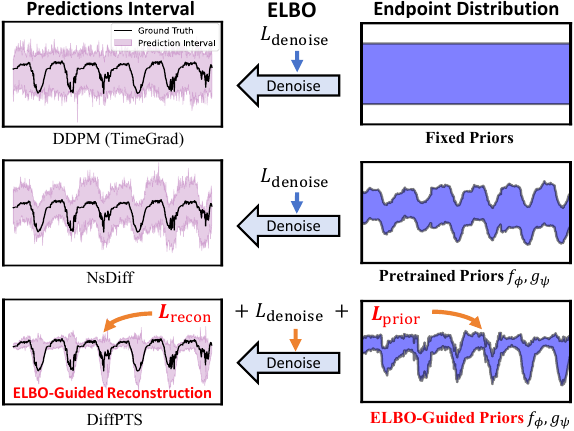}
  \caption{Figure~\ref{fig:introduction} illustrates the comparison of DDPM-based models on the ETTm1 dataset. The left panel displays the prediction intervals, while the right panel shows the corresponding endpoint distributions for TimeGrad, NsDiff, and DiffPTS (ours), respectively. Both TimeGrad and NsDiff utilize the standard ELBO objective from DDPM, whereas DiffPTS reformulates the ELBO to enable end-to-end training within a unified framework.}
  \label{fig:introduction}
\end{wrapfigure}

Figure~\ref{fig:introduction} presents a comparative analysis on the ETTm1 (Electricity Transformer Temperature, minutes) dataset, showcasing estimated prediction intervals (Left) and endpoint distributions (Right) across three diffusion-based models: TimeGrad~\cite{rasul2021autoregressive}, NsDiff~\cite{ye2025non}, and our proposed DiffPTS.  TimeGrad~\cite{rasul2021autoregressive} follows the conventional DDPM framework with a fixed Gaussian prior, which fails to capture the temporal variability inherent in time series data.  While NsDiff allows for dynamic mean and variance estimation by introducing the adaptive endpoint $\mathcal{N}(f(\mathbf{X}), g(\mathbf{X}))$, $f_\phi,g_\psi$ are considered as priors pretrained with designed losses that may not be inherently aligned with the diffusion process; as a result, the sliding variance target for $g_\psi$ leads to an over-estimation in the uncertainty.

Based on this observation, we rethink the ELBO under the Location-Scale Noise Model (LSNM) with estimators $f_\phi, g_\psi$, and find that all three terms (reconstruction, denoise, prior matching) in the ELBO inherently require strict optimization, whereas two of them are largely discarded in recent methods. By reformulating the ELBO to explicitly incorporate these components, we establish a unified optimization objective that seamlessly integrates the training of estimators with the core diffusion training process. As in Figure~\ref{fig:introduction}, DiffPTS optimizes these key components under the guidance of the core ELBO that is inherently required by the diffusion process, leading to a more accurate estimation~(More explanation is given in Figure~\ref{fig:sample_showcase}).

In summary, our contributions are: 1) We identify a critical theoretical puzzle in recent advances for diffusion-based probabilistic forecasting: while trainable parameters are introduced to model distributional shift, they are not optimized within the diffusion ELBO framework. 2) We reformulate the diffusion ELBO to properly incorporate the introduced mean and variance estimators. Our analysis reveals that the ELBO inherently specifies a Gaussian negative log likelihood (NLL) objective for both estimators, indicating that prior approaches relying on manually designed targets may be fundamentally misaligned with the diffusion process. Moreover, we show that the ELBO provides a unified objective for both residual-based and endpoint-based methods. 3)  Based on the reformulated ELBO, we propose DiffPTS, a novel end-to-end diffusion framework for PTSF. Extensive experiments demonstrate that DiffPTS consistently yields significant improvements in probabilistic forecasting performance. Notably, it outperforms previous diffusion-based baselines by over 14.53\%/16.55\% in CRPS/MSE across nine benchmarks, highlighting the importance of a full ELBO formulation.

\section{Background}


\subsection{Denoising Diffusion Probabilistic Model}
Diffusion probabilistic models~\cite{sohl2015deep} define a generative process by modeling the data distribution $p(\mathbf{Y}_0)$ as the marginal of a joint distribution over a sequence of latent variables $\mathbf{Y}_{1}, \dots, \mathbf{Y}_{T}$: $p(\mathbf{Y}_0) := \int p(\mathbf{Y}_{0:T}) \, d\mathbf{Y}_{1:T}$. Denoising Diffusion Probabilistic Models (DDPMs)~\cite{ho2020denoising} construct the diffusion process based on a Markovian chain~\cite{chung1967markov}: $p(\mathbf{Y}_{0:T}) := p(\mathbf{Y}_T) \prod_{t=1}^T p(\mathbf{Y}_{t-1} \mid \mathbf{Y}_t)$. In DDPM, the endpoint distribution is fixed to a Gaussian prior, i.e., $p(\mathbf{Y}_T) = \mathcal{N}(\mathbf{0}, \mathbf{I})$. The model comprises two complementary processes: a \emph{forward} process that progressively corrupts data by adding Gaussian noise, and a \emph{reverse} (or \emph{denoising}) process that learns to recover the original data by iteratively removing noise~\cite{nakkiran2025step}. The forward process is defined as:
\begin{align}
q(\mathbf{Y}_{1:T} \mid \mathbf{Y}_0) &:= \prod_{t=1}^T q(\mathbf{Y}_t \mid \mathbf{Y}_{t-1}), \quad q(\mathbf{Y}_t \mid \mathbf{Y}_{t-1}) := \mathcal{N}\big(\mathbf{Y}_t; \sqrt{1 - \beta_t} \, \mathbf{Y}_{t-1}, \, \beta_t \mathbf{I} \big),
\label{eq:diff_paradigm_forward_conventional}
\end{align}
where $\beta_t \in (0,1) $ is a pre-specified noise schedule that controls the variance of added noise at each step. The reverse process involves denoising $\mathbf{Y}_T$ back to $\mathbf{Y}_0$  and is parameterized by a learnable Gaussian transition kernel: $p_\theta(\mathbf{Y}_{t-1}|\mathbf{Y}_{t})$:
\begin{equation}
    p_\theta(\mathbf{Y}_{t-1}|\mathbf{Y}_{t}):=\mathcal{N}(\mathbf{Y}_{t-1};\boldsymbol{\mu}_\theta(\mathbf{Y}_t, t), \boldsymbol{\sigma}_\theta(\mathbf{Y}_t,t)),
\end{equation}
\begin{equation}
 \boldsymbol{\mu}_\theta(\mathbf{Y}_t, t) := \frac{1}{\sqrt{\alpha_t}}(\mathbf{Y}_t-\frac{\beta_t}{\sqrt{1-\bar\alpha_t}}\boldsymbol{\epsilon}_\theta(\mathbf{Y}_t,t)) ,\quad \boldsymbol{\sigma}_\theta(\mathbf{Y}_t,t) = \frac{1-\bar\alpha_{t-1}}{1-\bar\alpha_t}\beta_t\mathbf{I},
\end{equation}
where $\alpha_t := 1 - \beta_t$, $\bar{\alpha}_t := \prod_{i=1}^t \alpha_i$, and $ \boldsymbol{\epsilon}_\theta(\cdot, t) $ denotes a neural network trained to predict the noise added at step $t$.  Training in DDPM is performed by maximizing a variational lower bound on the log-likelihood, known as the Evidence Lower Bound (ELBO)~\cite{lee2021priorgrad,luo2022understanding}:
\begin{align}
\text{ELBO}&=\; \mathbb{E}_{q(\mathbf{Y}_1 \mid \mathbf{Y}_0)} \left[ \log p_\theta(\mathbf{Y}_0 \mid \mathbf{Y}_1) \right] \nonumber   - D_{\text{KL}}\big( q(\mathbf{Y}_T \mid \mathbf{Y}_0) \,\|\, p(\mathbf{Y}_T) \big) \\
&- \sum_{t=2}^{T} \mathbb{E}_{q(\mathbf{Y}_t \mid \mathbf{Y}_0)} \left[ D_{\text{KL}}\big( q(\mathbf{Y}_{t-1} \mid \mathbf{Y}_t, \mathbf{Y}_0) \,\|\, p_\theta(\mathbf{Y}_{t-1} \mid \mathbf{Y}_t) \big) \right].
\label{eq:elbo_standard}
\end{align}
The first, second, and third terms in the ELBO objective are commonly referred to as the \textit{reconstruction},  \textit{prior matching}, and \textit{denoising} terms, respectively. Crucially, in standard DDPM, the prior matching term is a KL divergence between two fixed distributions, which can be analytically computed and thus ignored during optimization~\cite{kingma2013auto, goodfellow2020generative}. Similarly, the reconstruction term involving $q(\mathbf{Y}_0|\mathbf{Y}_1)$ corresponds to a fixed distribution and can be reduced to the denoising term at $t=1$.
 
As noted in~\cite{ho2020denoising}, $p_\theta(\cdot)$ can be reparameterized such that the dominant contribution to the loss arises from predicting the noise injected during the forward process. Consequently, the training objective simplifies to minimizing the following denoising loss:
\begin{align}
\min_\theta \mathcal{L}_{\text{denoise}}(\theta) = \mathbb{E}_{\mathbf{Y}_0 \sim q(\mathbf{Y}_0),\, \boldsymbol{\epsilon} \sim \mathcal{N}(\mathbf{0}, \mathbf{I}),\, t \sim \mathcal{U}\{1,\dots,T\}} \Big[ \big\| \boldsymbol{\epsilon}_0 - \boldsymbol{\epsilon}_\theta(\mathbf{Y}_t, t) \big\|^2 \Big],
\label{eq:ddpm_denoise_loss}
\end{align}
where $\boldsymbol{\epsilon}_0 \sim \mathcal{N}(\mathbf{0}, \mathbf{I})$ and $\mathbf{Y}_t = \sqrt{\bar{\alpha}_t} \, \mathbf{Y}_0 + \sqrt{1 - \bar{\alpha}_t} \, \boldsymbol{\epsilon}_0$ is sampled directly via the reparameterization trick. At inference time, new samples are generated by first drawing $ \mathbf{Y}_T \sim \mathcal{N}(\mathbf{0}, \mathbf{I}) $ and then iteratively applying the learned reverse process from $ t = T $ down to $ t = 1 $.

\subsection{Probabilistic Time Series Forecasting}
Given a historical multivariate time series sample $\mathbf{X} \in \mathbb{R}^{N \times D}$, where $N$ denotes the look-back window length and $D$ the number of variables, the goal of probabilistic forecasting is to characterize the conditional distribution of the future trajectory $p(\mathbf{Y})=\{p(\mathbf{y}_1), p(\mathbf{y}_2), ..., p(\mathbf{y}_M)|\mathbf{y}\in\mathbb{R}^{D}\}$, where $M$ is the future window size. In this context, this DDPM framework is naturally extended to \emph{conditional} DDPM~\cite{choi2021ilvr}, which incorporates past observations $\mathbf{X}$ as an additional conditioning variable, and the conditional DDPM is formulated as:
\begin{equation}
p_\theta(\mathbf{Y}_{0:T}|\mathbf{X}) := p(\mathbf{Y}_T|\mathbf{X}) \prod_{t=1}^T p_\theta(\mathbf{Y}_{t-1} \mid \mathbf{Y}_t, \mathbf{X}),
\end{equation}
\begin{equation}
p_\theta(\mathbf{Y}_{t-1} \mid \mathbf{Y}_t, \mathbf{X}) :=  \mathcal{N}(\mathbf{Y}_{t-1};\boldsymbol{\mu}_\theta(\mathbf{Y}_t, \mathbf{X}, t ), \boldsymbol{\sigma}_\theta(\mathbf{Y}_t, \mathbf{X},t)).
\end{equation}
The reverse transition is typically parameterized by incorporating an additional conditioning input to the denoising network: $ \boldsymbol{\epsilon}_\theta(\cdot, t; \mathbf{X}) $. Beyond this parameterization, several studies~\cite{tashiro2021csdi,rasul2021autoregressive, shen2023non} have proposed mechanisms to incorporate various conditional information, e.g., mixed historical codes, into the reverse process to better guide the denoising function. Notably, both the ELBO (Equation~\eqref{eq:elbo_standard}) and the simplified denoising loss (Equation~\eqref{eq:ddpm_denoise_loss}) retain their original form in these methods.

Nevertheless, PTSF poses unique challenges  due to strong temporal dependencies~\cite{wu2021autoformer,wu2022timesnet,zhou2021informer}, non-stationarity~\cite{kim2021reversible, liu2024timebridge}, and input-dependent uncertainty~\cite{nakkiran2025step, salinas2020deepar}. 
While diffusion models have shown promise in this domain~\cite{tashiro2021csdi,rasul2021autoregressive}, the conventional paradigm, diffusing raw observations $\mathbf{Y}$ toward a standard Gaussian prior, often overlooks the input-dependent nature of predictive uncertainty~\cite{dai2025samples,lu2026towards}.

To better align with the forecasting task, recent works propose to reformulate the diffusion process by leveraging a deterministic predictor $f_\phi(\mathbf{X})$ and uncertainty estimator $g_\psi(\mathbf{X})$ trained on the context $\mathbf{X}$. 
For instance, some methods diffuse the residual $(\mathbf{Y} - f_\phi(\mathbf{X}))/g_\psi(\mathbf{X})$ instead of $\mathbf{Y}$ itself~\cite{li2025diffusion,lai2025rdit}, while others replace the standard prior with an adaptive endpoint $\mathcal{N}(f_\phi(\mathbf{X}), g_\psi(\mathbf{X}))$~\cite{ye2025non,li2024tmdm}. 
These designs implicitly assume that the diffusion endpoint $\mathbf{Y}_T$ follows a \textit{Location-scale noise model} (LSNM)~\cite{ye2025non}, where both the mean and uncertainty are input-dependent:
\begin{equation}
\label{eq:LSNM}
    \mathbf{Y}_T = f_\phi(\mathbf{X}) + \sqrt{g_\psi(\mathbf{X})} \odot \boldsymbol{\eta},
\end{equation}
where $\boldsymbol{\eta} \sim \mathcal{N}(\mathbf{0}, \mathbf{I})$, and $\odot$ denotes element-wise multiplication (all operations are element-wise unless indicated). Here, $f_\phi(\mathbf{X}) \in  \mathbb{R}^{M \times D}$ estimates the conditional mean $\mathbb{E}[\mathbf{Y} \mid \mathbf{X}]$, while $g_\psi(\mathbf{X}) \in \mathbb{R}_{>0}^{M \times D}$ models input-dependent uncertainty, both can be interpreted as prior without parameters.   

\section{Discussion \& Methodology}
\label{sec:method}
In this section, we rethink the foundational ELBO of recent diffusion-based methods for probabilistic time series forecasting.  We summarize existing approaches and find that, despite introducing auxiliary estimators, these components are not dictated by the standard ELBO and are instead added heuristically.  Based on this observation, we argue that a proper variational objective should explicitly govern the form and role of these estimators.  To this end, we derive a corrected evidence lower bound that imposes precise requirements on the introduced estimators, ensuring a tighter variational bound.  Building on this formulation, we present \textbf{DiffPTS}, a novel probabilistic forecasting framework trained in an end-to-end joint optimization scheme and we describe its  training and sampling algorithms.

\label{headings}


\subsection{The Missing Puzzle of The ELBO}

Recent diffusion-based approaches for probabilistic time series forecasting have achieved notable success. To motivate our work, we first summarize the design choices of current state-of-the-art DDPM frameworks, specifically their denoising targets, reverse endpoints, and associated diffusion objectives under the LSNM assumption, in the following table:
\begin{table}[h]
\centering
\caption{Summary of recent diffusion-based advancements for probabilistic time series forecasting. We consider the denoising target and reverse endpoint. $\to$ indicates a supervised objective and $+$ represents an additional objective compared to previous methods. $\mathcal{N}(f_\phi(\mathbf{X}),\, g_\psi(\mathbf{X}))$ denotes $\mathcal{N}(f_\phi(\mathbf{X}),\, \mathrm{diag}(g_\psi(\mathbf{X})))$; we omit $\mathrm{diag}(\cdot)$ for clarity as only diagonal variance is modeled. RDIT uses a  training residual $g_\psi=\boldsymbol{\sigma}_{trn}$, and NsDiff estimates $\boldsymbol{\sigma}_\mathbf{Y}$ using future sliding window variance.}
\setlength{\tabcolsep}{9pt}
\begin{tabular}{lccc}
\toprule
\textbf{Models} & \textbf{Target} & \textbf{Endpoint} & \textbf{Objectives} \\
\midrule
DDPM (\cite{tashiro2021csdi}, \cite{rasul2021autoregressive}, \cite{shen2023non}) 
    & $ \mathbf{Y}$ & $\mathcal{N}(\mathbf{0}, \mathbf{I})$ & $(\epsilon_\theta \to \epsilon_0)$ \\
TMDM \cite{li2024tmdm}                 
    & $\mathbf{Y}$ & $\mathcal{N}(f_{\phi}(\mathbf{X}), \mathbf{I})$ & \multirow{2}{*}{$+( f_\phi(\mathbf{X}) \to \mathbf{Y})$} \\
D3U \cite{li2025diffusion}                  
    & $\mathbf{Y} - f_{\phi}(\mathbf{X})$ & $\mathcal{N}(\mathbf{0},\mathbf{I})$ & \\
RDIT \cite{lai2025rdit}                  
    & $(\mathbf{Y} - f_{\phi}(\mathbf{X})) / \boldsymbol{\sigma}_{trn}$ & $\mathcal{N}(\mathbf{0}, \mathbf{I})$ & \multirow{2}{*}{$+(g_\psi(\mathbf{X}) \to \sigma_{\mathbf{Y}})$} \\
NsDiff \cite{ye2025non}               
    & $\mathbf{Y}$ & $\mathcal{N}(f_{\phi}(\mathbf{X}), g_\psi({\mathbf{X}}))$ & \\
\bottomrule
\label{tab:method_comparison}
\end{tabular}
\end{table}

As summarized in Table~\ref{tab:method_comparison}, recent advances share a common design principle: they depart from the standard paradigm of diffusing raw observations $\mathbf{Y}$ by introducing  residual modeling~\cite{li2025diffusion}, variance normalization~\cite{lai2025rdit}, or context-dependent priors~\cite{ye2025non,li2024tmdm} into the generative pipeline. 
To realize these designs, all existing methods require \textit{additional estimators} beyond the core denoising network $\epsilon_\theta$, i.e., a deterministic forecaster $f_\phi(\mathbf{X})$ or a variance estimator $g_\psi(\mathbf{X})$.


However, a critical question remains largely unexamined: \textit{when these methods introduce external components (e.g., a base forecaster $f_\phi$ or a variance estimator $g_\psi$), are these components theoretically grounded in the variational objective, or are they heuristically appended without proper alignment to the ELBO?} Notably, despite their empirical effectiveness, none of these methods derive a revised evidence lower bound that accounts for the input-dependent transformation implied by the LSNM. Consequently, the auxiliary estimators are trained via ad-hoc supervised losses (e.g., MSE for $f_\phi$ or sliding-window variance for $g_\psi$) that are decoupled from the diffusion likelihood objective.

\subsection{A Unified Formulation of The ELBO}
To address this theoretical puzzle, we revisit the ELBO by  considering the learnable parameters $\phi, \psi$ that are introduced in these methods. Without loss of generality, we consider the general case where the reverse endpoint distribution is $\mathcal{N}(f_{\phi}(\mathbf{X}), g_\psi(\mathbf{X}))$. Previous work~\cite{ye2025non} considers an uncertainty-aware diffusion schedule, requiring prior knowledge of the data’s variance structure. To achieve a unified formulation, we adopt the following forward process with a standard noise schedule:
\begin{equation}
\mathbf{Y}_t=\sqrt{\bar{\alpha}_t} \mathbf{Y}_0+\left(1-\sqrt{\bar{\alpha}_t}\right) f_\phi(\mathbf{X})+\sqrt{\left(1-\bar{\alpha}_t\right) g_\psi(\mathbf{X})}\boldsymbol{\epsilon}_0,
\label{eq:forward_Yt_Nsdiff}
\end{equation}
the forward process admits a reverse posterior as~\cite{ye2025non, li2024tmdm, han2022card}:
\begin{equation}
q(\mathbf{Y}_{t-1} \mid \mathbf{Y}_t, \mathbf{Y}_0, \mathbf{X}) = \mathcal{N}\left( \tilde{\boldsymbol{\mu}}_t(\mathbf{Y}_t, \mathbf{Y}_0, \mathbf{X}),\;\tilde{\beta}_t g_\psi(\mathbf{X}) \right) \label{eq:forward_posterior}
\end{equation}
\begin{equation}
\tilde{\boldsymbol{\mu}}_t:=\frac{\beta_t \sqrt{\bar{\alpha}_{t-1}}}{1-\bar{\alpha}_t} \mathbf{Y}_0+\frac{\left(1-\bar{\alpha}_{t-1}\right) \sqrt{\alpha_t}}{1-\bar{\alpha}_t} \mathbf{Y}_t+\left(1+\frac{\left(\sqrt{\bar{\alpha}_t}-1\right)\left(\sqrt{\alpha_t}+\sqrt{\bar{\alpha}_{t-1}}\right)}{1-\bar{\alpha}_t}\right) f_\phi(\mathbf{X}),
\label{eq:forward_posterior_mean}
\end{equation}
where $\tilde{\beta}_t = \frac{\bar{\beta}_{t-1}}{\bar{\beta}_t} \beta_t$. We leave the derivation in Appendix~\ref{sec:proof:proposition:complete_elbo}.  By parameterizing a posterior approximator $p_\theta(\mathbf{Y}_{t-1} \mid \mathbf{Y}_t, \mathbf{X})$,  the ELBO on three parameters $\phi,\psi,\theta$ is derived as follows:


\begin{proposition}
Let $\boldsymbol{\epsilon}_0 \sim \mathcal{N}(\mathbf{0}, \mathbf{I})$ and $(\mathbf{X}, \mathbf{Y}_0) \sim q_{\text{data}}$. Under the forward process defined in Equation~\eqref{eq:forward_Yt_Nsdiff} and with the prior $p(\mathbf{Y}_T \mid \mathbf{X}) = \mathcal{N}(f_\phi(\mathbf{X}), g_\psi(\mathbf{X}))$, the ELBO can be expressed as:
\begin{equation}
-\mathrm{ELBO} = C + \sum_{t=1}^{T} \gamma_t \mathbb{E}_{\boldsymbol{\epsilon}_0} \left[ \| \boldsymbol{\epsilon}_0 - \boldsymbol{\epsilon}_\theta(\mathbf{Y}_t, \mathbf{X}, t) \|^2 \right] + \frac{\bar{\alpha}_T}{2}\|\frac{ f_\phi(\mathbf{X}) - \mathbf{Y}_0 }{\sqrt{g_\psi(\mathbf{X})}} \|^2 + \sum_{i=1}^d\frac{\log \left[g_\psi(\mathbf{X})\right]_i}{2} 
\label{eq:complete_elbo_maintext}
\end{equation}
where C is a constant independent of $\theta, \phi$ and $\psi$, $\gamma_t=\frac{\beta_t}{2 \alpha_t\left(1-\bar{\alpha}_{t-1}\right)} \text { for } t>1, \text { and } \gamma_1=\frac{1}{2 \alpha_1}$.
\label{proposition:complete_elbo}
\end{proposition}




A complete derivation is provided in Appendix~\ref{sec:proof:proposition:complete_elbo}. Proposition~\ref{proposition:complete_elbo} presents a parameterized extension of the standard ELBO in Equation~\eqref{eq:elbo_standard} by introducing two learnable functions $f_\phi(\mathbf{X})$ and $g_\psi(\mathbf{X})$.  Notably, despite differences in their original formulations, such as whether they train on residual targets (e.g., RDIT) or adaptive endpoints (e.g., NsDiff), all these methods can be unified under this ELBO. This leads to the following key insight:



\begin{remark}
Under a shared noise schedule, the ELBOs of NsDiff, RDIT, D3U, TMDM, and DDPM are structurally equivalent and reduce to Equation~\eqref{eq:complete_elbo_maintext} under appropriate instantiations of $f_\phi, g_\psi$:
\begin{equation}
\mathrm{ELBO}_{\mathrm{NsDiff}} \xrightarrow{g_\psi = \boldsymbol{\sigma}_{trn}} \mathrm{ELBO}_{\mathrm{RDIT}} \xrightarrow{g_\psi = 1} \mathrm{ELBO}_{\mathrm{D3U, TMDM}} \xrightarrow{f_\phi = 0}\mathrm{ELBO}_{\mathrm{DDPM}}
\end{equation}
\label{remark:model_equivalence}
\end{remark}
The proof of this equivalence is detailed in Appendix~\ref{sec:proof:model_equivalence}. This unification  reveals that: despite their differing diffusion paradigms, all these approaches optimize the same ELBO objective under a standard denoising schedule. Consequently, the ELBO can be seamlessly applied to any of these frameworks to achieve a tighter ELBO bound and thereby improve generative and predictive performance. We provide plug-and-play results in Section~\ref{subsec:plugin_experiment}.

\subsection{A General Framework: DiffPTS}
\begin{figure}[htbp]
\includegraphics[width=1\linewidth]{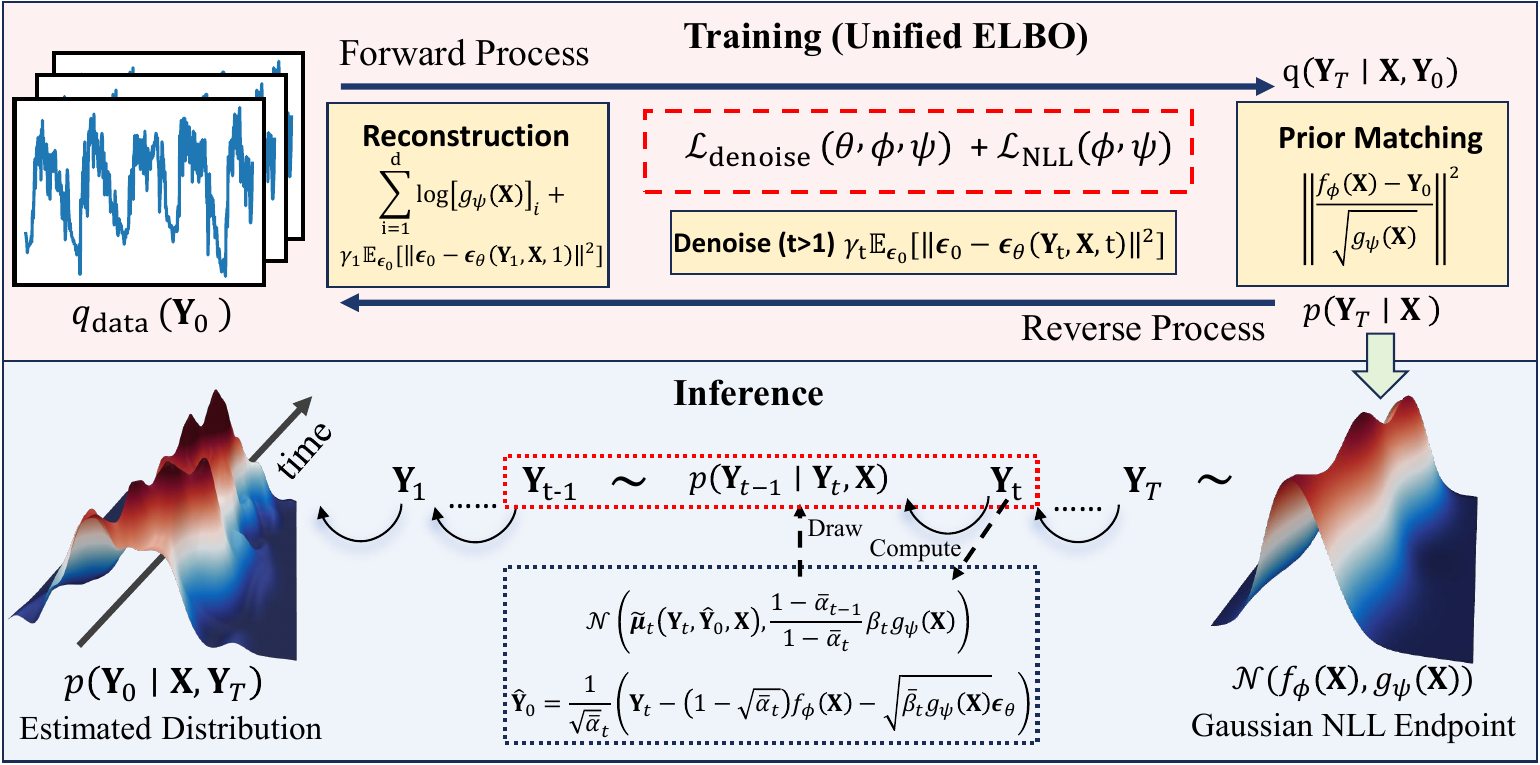}
\caption{Overview of the proposed DiffPTS framework. Training (Top): We optimize the proposed unified ELBO, encompassing: Reconstruction, Denoising, and Prior Matching. This training procedure is paradigm-agnostic and can be readily adapted to other methods in Table~\ref{tab:method_comparison}. 
Inference (Bottom): The generative process initiates by sampling from the learned NLL endpoint. The model then iteratively samples from the reverse distribution $p_\theta(\mathbf{Y}_{t-1}|\mathbf{Y}_t, \mathbf{X})$ to progressively recover the complex target distribution $p(\mathbf{Y}_0|\mathbf{X}, \mathbf{Y}_T)$. All visualizations are based on the ETTm1 dataset. }
\label{fig:overview}
\end{figure}
Building upon the generalized ELBO derived in Proposition~\ref{proposition:complete_elbo}, we now instantiate our proposed framework, DiffPTS. We describe its objective function, training, and sampling algorithms. DiffPTS offers a generalized approach for the diffusion-based model without pretrained estimators. We present an overview of the proposed framework in Figure~\ref{fig:overview}. Specifically, DiffPTS relies on the proposed ELBO in the training phase, covering reconstruction, denoising, and prior matching. Compared to previous works introducing additional parameters by pretraining, our objective can be interpreted as jointly training the denoising process and a Gaussian NLL endpoint, as described below:

\begin{proposition}\label{prop:loss_function_DiffPTS}
The negative ELBO minimization $\mathcal{L}_{\text{ELBO}}(\theta, \phi, \psi)$ reduces to the minimization of $\mathcal{L}_{\text{denoise}}(\theta, \phi, \psi)$ and a Gaussian negative log likelihood $\mathcal{L}_{\text{NLL}}(\phi, \psi)$: 
\begin{equation}
\mathcal{L}_{\text {ELBO}}(\theta, \phi, \psi)= \underbrace{\mathbb{E}_{t, \boldsymbol{\epsilon}_0} \left[ \| \boldsymbol{\epsilon}_0 - \boldsymbol{\epsilon}_\theta(\mathbf{Y}_t, \mathbf{X}, t) \|^2 \right] \vphantom{\frac{1}{2}
\left\|\frac{ f_\phi(\mathbf{X}) - \mathbf{Y}_0 }{\sqrt{g_\psi(\mathbf{X})}}\right\|^2+\sum_{i=1}^d\frac{\log \left[g_\psi(\mathbf{X})\right]_i}{2}}
}_{\mathcal{L}_{\text{denoise}}(\theta, \phi, \psi)} + \underbrace{\frac{1}{2}\|\frac{ f_\phi(\mathbf{X}) - \mathbf{Y}_0 }{\sqrt{g_\psi(\mathbf{X})}} \|^2 + \sum_{i=1}^d\frac{\log \left[g_\psi(\mathbf{X})\right]_i}{2}}_{\mathcal{L}_{\text{NLL}}(\phi, \psi)}
\label{eq:denoise_proposition_diffpts}
\end{equation}

\end{proposition}

We leave the proof in Appendix~\ref{apdx:proof:prop:loss_function_DiffPTS}. This decomposition offers an intuitive and principled learning paradigm: \textit{we initialize the generative process from a tractable conditional Gaussian NLL prior parameterized by $(f_\phi(\mathbf{X}), g_\psi(\mathbf{X}))$, and then refine it through a diffusion-based denoising procedure to capture complex, non-Gaussian structure in the target distribution $p(\mathbf{Y} \mid \mathbf{X})$}. \textbf{Particularly, the training of $f_\phi$ and $g_\psi$ is not arbitrarily specified but should be strictly derived from the Gaussian NLL objective}, in contrast to prior works that rely on manually designed loss functions and two-phase training paradigms which may be suboptimal~\cite{ye2025non, li2024tmdm}. Importantly, although $\mathcal{L}_{\text{denoise}}$ is formally similar to the denoising loss in Equation~\eqref{eq:ddpm_denoise_loss}, it actively contributes to optimizing the endpoint parameters $(\phi, \psi)$, as $\mathbf{Y}_t$ in Equation~\eqref{eq:forward_Yt_Nsdiff} explicitly depends on these parameters. 

When the variance estimator is fixed to constant variance, i.e., $g_\psi(\mathbf{X}) \equiv \mathbf{1}$, the Gaussian NLL term reduces to  a mean squared error~(MSE) $\| f_\phi(\mathbf{X}) - \mathbf{Y}_0 \|^2$ which recovers the MSE explicitly added in D3U~\cite{li2025diffusion} and TMDM~\cite{li2024tmdm} for training their conditional predictors. Furthermore, unlike recent methods such as NsDiff~\cite{ye2025non} and RDIT~\cite{lai2025rdit}, which rely on auxiliary objectives or hand-crafted rules to estimate the conditional variance $g_\psi(\mathbf{X})$, \textit{our approach learns $g_\psi(\mathbf{X})$ implicitly without a supervised target and optimally achieves a tighter ELBO}. 

Algorithms~\ref{alg:diffpts_training} and ~\ref{alg:diffpts_sampling} describe the training and sampling procedures of the proposed DiffPTS. Algorithm~\ref{alg:diffpts_training} implements an end-to-end training paradigm that jointly optimizes the reverse posterior $p_\theta(\mathbf{Y}_{t-1}|\mathbf{Y}_t, \mathbf{X})$ and a Gaussian prior $\mathcal{N}(f_\phi(\mathbf{X}), g_\psi(\mathbf{X}))$ through the proposed ELBO formulation.  Crucially, this formulation eliminates the need for pretraining and hand-crafted variance supervision by directly modeling the data's intrinsic uncertainty in previous works~\cite{ye2025non, li2025diffusion, lai2025rdit, li2024tmdm, han2022card}. During sampling (Algorithm~\ref{alg:diffpts_sampling}), the generative process begins from the learned prior distribution $\mathcal{N}(f_\phi(\mathbf{X}), g_\psi(\mathbf{X}))$ and then iteratively generates $\mathbf{Y}_{t-1}$ by sampling from $p_\theta(\mathbf{Y}_{t-1}|\mathbf{Y}_t, \mathbf{X}) \approx q(\mathbf{Y}_{t-1} \mid \mathbf{Y}_t, \hat{\mathbf{Y}}_0, \mathbf{X})$ similar to  Equation~\eqref{eq:forward_posterior}~\cite{ye2025non, li2024tmdm, han2022card}.
\noindent\begin{minipage}[t]{0.4\linewidth}
\begin{algorithm}[H]
   \caption{Training Procedure}
   \label{alg:diffpts_training}
\begin{algorithmic}[1]
   \STATE {\bfseries Input:} $\mathbf{X}, \mathbf{Y}_0, f_\phi, g_\psi, \epsilon_\theta, T$
   \REPEAT
   \STATE Draw $\boldsymbol{\epsilon}_0 \sim \mathcal{N}(\mathbf{0},\mathbf{I})$
   \STATE Draw $t \sim \mathcal{U}(\{1, \dots, T\})$
   \STATE Compute $\mathbf{Y}_t = \sqrt{\bar{\alpha}_t} \mathbf{Y}_0 + (1 - \sqrt{\bar{\alpha}_t}) f_{\phi}(\mathbf{X}) + \sqrt{ \bar\beta_t g_\psi(\mathbf{X}) } \boldsymbol{\epsilon}_0$
   \STATE $\boldsymbol{\epsilon}_\theta \gets \epsilon_\theta(\mathbf{Y}_t, \mathbf{X}, t)$
   \STATE Compute $\mathcal{L}_{\text{ELBO}}$ \hfill $\rhd$ Equation~\ref{eq:denoise_proposition_diffpts}
   \STATE Update $(\theta, \phi, \psi)$ via $\nabla \mathcal{L}_{\text{ELBO}}$
   \UNTIL convergence
\vspace{0.39em}
\end{algorithmic}
\end{algorithm}
\end{minipage}%
\begin{minipage}[t]{0.6\linewidth}
\begin{algorithm}[H]
   \caption{Sampling Procedure}
   \label{alg:diffpts_sampling}
\begin{algorithmic}[1]
   \STATE {\bfseries Input:} $\mathbf{X}, f_\phi, g_\psi, \epsilon_\theta, T$
   \STATE Draw $\mathbf{Y}_T \sim \mathcal{N}(f_\phi(\mathbf{X}), g_\psi(\mathbf{X})), \boldsymbol{z} \sim \mathcal{N}(\mathbf{0}, \mathbf{I})$
   \FOR{$t = T, \dots, 1$}
   \STATE $\boldsymbol{\epsilon}_\theta \gets \epsilon_\theta(\mathbf{Y}_t, \mathbf{X}, t)$
   \STATE $ \hat{\mathbf{Y}}_0 \gets \frac{1}{\sqrt{\bar{\alpha}_t}} (
   \mathbf{Y}_t 
   - \left(1 - \sqrt{\bar{\alpha}_t}\right) f_{\phi}(\mathbf{X}) 
   - \sqrt{ \bar\beta_tg_\psi(\mathbf{X})}\boldsymbol{\epsilon}_\theta
)$
\STATE $\mathbf{Y}_{t-1} \gets 
    \begin{cases}
        \tilde{\boldsymbol{\mu}}_t(\mathbf{Y}_t, \hat{\mathbf{Y}}_0, \mathbf{X}) + \sqrt{\tilde\beta_t\, g_\psi(\mathbf{X})}\, \boldsymbol{z}, \ t > 1 \\
        \hat{\mathbf{Y}}_0, \quad \quad  \quad  \quad  \quad  \quad  \quad  \quad  \quad  \quad  \quad  \quad t=1
    \end{cases}$
   \ENDFOR
   \STATE \textbf{return} $\mathbf{Y}_0$
\end{algorithmic}
\end{algorithm}
\end{minipage}

\section{Experiments}
\label{sec:Experiments}

\subsection{Experiment Setup}
\label{subsec:exp_setup}
\textbf{Datasets}. Following previous work~\cite{ye2025non,li2024tmdm}, we evaluate on nine real-world datasets: ETT \{h1, h2, m1, m2\}, Electricity (ECL), Influenza-like Illness (ILI), Exchange Rate (EXR), Traffic, and Solar Energy (Solar). All experiments adopt a input sequence length of 168 time steps, ILI uses 32 prediction steps, and other datasets follow 192 prediction steps.  Data splits follow domain conventions: ETT datasets use 12/4/4-month partitions for training, validation, and testing; all other datasets employ a chronological 7:1:2 split ratio. We provide dataset details in Appendix~\ref{apdx:subsec:dataset_details}.

\textbf{Baselines}. We select eight recent strong PTSF baselines for comparison, including TimeGrad~\cite{rasul2021autoregressive}, CSDI~\cite{tashiro2021csdi}, TimeDiff~\cite{shen2023non}, TMDM~\cite{li2024tmdm}, DiffusionTS~\cite{yuan2024diffusion}, D3U~\cite{li2025diffusion}, RDIT~\cite{lai2025rdit}, NsDiff~\cite{ye2025non}. Specifically, TimeGrad, CSDI, TimeDiff, DiffusionTS follow the standard diffusion paradigm; TMDM, D3U, RDIT, NsDiff introduce estimators into diffusion process through two phases pretraining.



\textbf{Implementation}. All experiments are conducted under popular long-term multivariate probabilistic forecasting settings~\cite{ye2025non,li2024tmdm}. We run all experiments with seeds $\{1, 2, 3\}$  for 10 epochs and select the best model based on the validation performance for final test evaluation. The learning rate is set to 0.001, batch size to 32, and the number of timesteps $T = 20$, consistent with prior work~\cite{ye2025non, rasul2021autoregressive}.  We employ a linear noise schedule with $\beta^1 = 10^{-4}$ and $\beta^T = 0.02$, in line with the setup used in conventional DDPM~\cite{ho2020denoising}. We use Adam optimizer~\cite{kingma2014adam} and the experiments were all performed using an NVIDIA RTX A6000 GPU. In inference, we generate 100 samples to estimate the distribution~\cite{desai2021timevae, ang2023tsgbench}. For the baseline models, we utilize their default parameters. For implementation of $f_\phi$ and $g_\psi$, we follow previous work~\cite{ye2025non, li2024tmdm} to use Non-stationary Transformer~\cite{liu2022non} and a three-layer linear model~\cite{popescu2009multilayer, gardner1998artificial}, respectively. More details are provided in Appendix~\ref{apdx:subsec:detail_implementation}.



\textbf{Metrics}. Following prior work's metric settings~\cite{ye2025non, rasul2021autoregressive, li2024tmdm}, we use Continuous Ranked Probability Score (CRPS)~\cite{matheson1976scoring} and mean square error~(MSE) to validate the performance of the estimated uncertainty quantification and point forecasting performance. For both metrics, smaller values indicate better performance. Detailed formula is provided in Appendix \ref{apdx:metrics}. 




\subsection{Main Experiments}
\label{subsec:exp_main}

The experimental results are presented in Table~\ref{tab:main_results}. DiffPTS demonstrates state-of-the-art performance across all nine real-world datasets in probabilistic time series forecasting, outperforming all competitive baselines including recent diffusion-based models such as NsDiff, RDIT, and D3U. On the ECL dataset, DiffPTS reduces CRPS by 24.8\% (from 0.290 to 0.218) and MSE by 13.9\% (from 0.209 to 0.180) compared to NsDiff. Similarly, on the Traffic dataset, DiffPTS achieves a 28.6\% reduction in CRPS (from 0.378 to 0.270) and a 28.4\% reduction in MSE (from 0.637 to 0.456). The results underscore the importance of optimizing the complete ELBO rather than relying on simplified denoising objectives. We provide results with two additional metrics in Appendix~\ref{subsec:exp_rst:main_exp}.


\begin{table}[htbp]
  \setstretch{0.9}
  \centering
  \footnotesize
  \caption{Experimental results on nine real-world datasets. \textbf{Bold face} indicates the best result, and \underline{underline} indicates the second-best result.}
    \begin{tabular}{ccccccccccc}
    \toprule
    Models & Datasets & ETTh1 & ETTh2 & ETTm1 & ETTm2 & ECL   & EXG   & ILI   & Solar & Traffic \\
    \midrule
    TimeGrad & CRPS  & 0.606  & 1.212  & 0.647  & 0.775  & 0.397  & 0.826  & 1.140  & 0.293  & 0.407  \\
    (2021) & MSE   & 1.062  & 3.462  & 1.218  & 1.690  & 0.505  & 1.567  & 4.197  & 0.475  & 0.983  \\
    \midrule
    CSDI  & CRPS  & 0.492  & 0.647  & 0.524  & 0.817  & 0.577  & 0.855  & 1.244  & 0.432  & 1.418  \\
    (2022) & MSE   & 0.949  & 1.226  & 1.002  & 1.723  & 1.007  & 1.701  & 4.515  & 0.763  & 1.731  \\
    \midrule
    TimeDiff & CRPS  & 0.465  & 0.471  & 0.464  & 0.316  & 0.750  & 0.433  & 1.153  & 0.700  & 0.771  \\
    (2023) & MSE   & \underline{0.517}  & \underline{0.456}  & 0.537  &  \underline{0.268}  & 0.879  & 0.402  & 3.958  & 0.821  & 1.350  \\
    \midrule
    DiffusionTS & CRPS  & 0.603  & 1.168  & 0.574  & 1.035  & 0.633  & 1.251  & 1.612  & 0.470  & 0.668  \\
    (2024) & MSE   & 1.089  & 3.273  & 1.030  & 2.372  & 1.072  & 3.628  & 6.053  & 0.749  & 1.473  \\
    \midrule
    TMDM  & CRPS  & 0.452  & 0.383  & 0.375  & 0.289  & 0.461  & 0.336  & 0.967  & 0.350  & 0.557  \\
    (2024) & MSE   & 0.696  & 0.512  & 0.494  & 0.315  & 0.257  & 0.334  & 3.636  & 0.250  & 0.679  \\
    \midrule
    D3U   & CRPS  & 0.447  & 0.385  & 0.412  & 0.282  & 0.439  & 0.354  & 2.097  & 0.488  & 0.507  \\
    (2025) & MSE   & 0.684  & 0.520  & 0.589  & 0.286  & 0.440  & 0.269  & 9.951  & 0.644  & 0.907  \\
    \midrule
    RDIT  & CRPS  & 0.410  & 0.380  & \underline{0.336}  & 0.258  & 0.362  & 0.392  & 0.845  & \underline{0.225}  & 0.495  \\
    (2025) & MSE   & 0.570  & 0.463  & 0.504  & 0.292  & 0.210  &  \underline{0.264}  & 2.857  &  \underline{0.180}  & 0.851  \\
    \midrule
    NsDiff & CRPS  & \underline{0.392}  &  \underline{0.358}  & 0.346  & \underline{0.256}  & \underline{0.290}  & \underline{0.324}  & \underline{0.806}  & 0.300  & \underline{0.378}  \\
    (2025) & MSE   & 0.594  & 0.514  &  \underline{0.488}  & 0.281  &  \underline{0.209}  & 0.300  &  \underline{2.846}  & 0.242  &  \underline{0.637}  \\
    \midrule
    DiffPTS & CRPS  & \textbf{0.374 } & \textbf{0.355 } & \textbf{0.329 } & \textbf{0.254 } & \textbf{0.218 } & \textbf{0.275 } & \textbf{0.731 } & \textbf{0.173 } & \textbf{0.270 } \\
    (ours) & MSE   & \textbf{0.507 } & \textbf{0.452 } & \textbf{0.424 } & \textbf{0.242 } & \textbf{0.180 } & \textbf{0.242 } & \textbf{2.636 } & \textbf{0.179 } & \textbf{0.456 }   \\ \bottomrule
    \end{tabular}%
  \label{tab:main_results}%
\end{table}%
\begin{figure}[ht]
\begin{center}
\centerline{\includegraphics[width=1\linewidth]{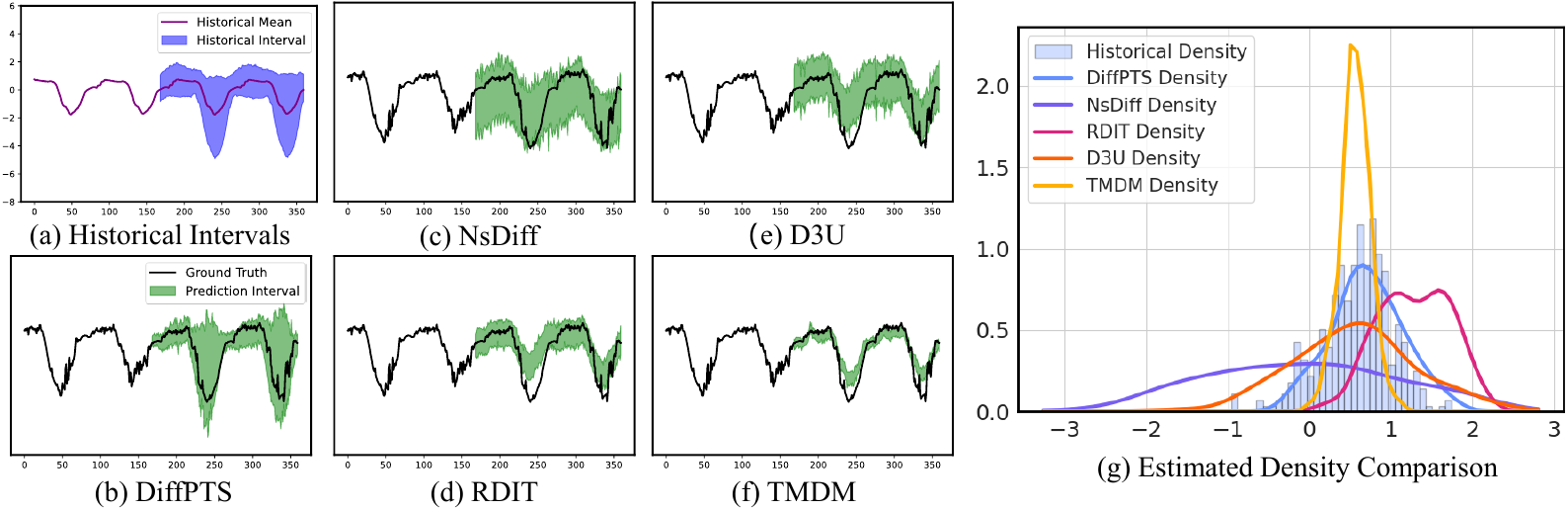}}
\caption{A visualization case of a ETTm1 sample. (a) The 96\% empirical distribution estimated on the samples with the same time. (b-f) The 96\%  prediction intervals of different methods. (g) Historical and estimated density of various methods, we use the last step of the same sample in (a-f).}
\label{fig:sample_showcase}
\end{center}
\vskip -0.3in
\end{figure}


\textbf{Sample Showcases}. To provide an illustration of DiffPTS's performance, we visualize 96\% distribution interval of a representative sample from the ETTm1 dataset in Figure~\ref{fig:sample_showcase}. And we compare to four recent state-of-the-art baselines: \textit{NsDiff}, \textit{RDIT}, \textit{D3U}, and \textit{TMDM}. Figure~\ref{fig:sample_showcase} (a) displays the historical ground-truth intervals derived from the mean-variance estimation of the same temporal sample.  As can be seen in Figure~\ref{fig:sample_showcase} (b), DiffPTS achieves remarkable consistency with this statistical estimate without relying on any explicit priors, effectively capturing both historical variance and mean information. Specifically, NsDiff (Figure~\ref{fig:sample_showcase} (c)) uses sliding window variance estimation, but remains incapable of capturing history-dependent variance patterns and thus cannot distinguish periodic information and variance; RDIT (Figure~\ref{fig:sample_showcase} (d)) employs training residuals for variance estimation, which amplifies errors when residual estimation is inaccurate; TMDM and D3U often neglect variance dynamics, leading to difficulties in achieving step-wise variance matching, as in Figure~\ref{fig:sample_showcase} (e-f), the models predict an inaccurate and fixed variance at each step.
This analysis underscores the importance of accurately modeling the ELBO. We provide additional showcases in Appendix~\ref{apdx:sec:other_showcases}.

\textbf{Distributional Analysis}. Beyond showcases, we compare the estimated densities of various methods against the historical ground truth distribution in Figure~\ref{fig:sample_showcase} (g). Our proposed DiffPTS (blue curve) demonstrates a superior alignment with the historical density (light blue). In contrast, both TMDM (yellow) and D3U (orange) show noticeable overestimation or underestimation of variance due to the fixed variance assumption; RDIT and NsDiff rely on hand-crafted supervised variance loss functions, which limits their ability to capture the true underlying distribution.  This precise distributional modeling capability underscores DiffPTS' effectiveness in learning time series distribution.



\subsection{Ablation Experiments}

To systematically investigate the effectiveness of each term in the proposed ELBO, we conduct ablation studies by removing loss components from the ELBO objective. Specifically, we evaluate three variants: (1) \textit{w/o NLL $f_\phi$}: excluding the $f_\phi$ optimization from Equation~\eqref{eq:complete_elbo_maintext} by setting $f_\phi(\mathbf{X}) = \mathbf{0}$; (2) \textit{w/o NLL $g_\psi$}: excluding the $g_\psi$ optimization from Equation~\eqref{eq:complete_elbo_maintext} by setting $g_\psi(\mathbf{X}) = \mathbf{1}$; and (3) \textit{w/o denoise}: excluding the denoising term from Equation~\eqref{eq:complete_elbo_maintext}. For sample generation, we utilize only the negative log-likelihood component, sampling directly from $\mathcal{N}(f_\phi(\mathbf{X}), g_\psi(\mathbf{X}))$.  The results, summarized in Table~\ref{tab:ablation_main}, demonstrate that the complete ELBO formulation achieves the best performance across all datasets. Notably, even when modeling only the ignored  terms (i.e., \textit{w/o denoise}), the model attains competitive performance, underscoring the critical importance of these terms in the diffusion process. Results with error bars are provided in Appendix~\ref{apdx:subsec:ablation_with_errorbar}.

\begin{table}[htbp]
  \centering
  \setstretch{0.9}
  \footnotesize
  \caption{Ablation results across nine datasets. \textbf{Bold} numbers indicate the best performance.}
    \begin{tabular}{ccccccccccc}
    \toprule
    Models & Datasets & \multicolumn{1}{c}{ETTh1} & \multicolumn{1}{c}{ETTh2} & \multicolumn{1}{c}{ETTm1} & \multicolumn{1}{c}{ETTm2} & \multicolumn{1}{c}{ECL} & \multicolumn{1}{c}{EXG} & \multicolumn{1}{c}{ILI} & \multicolumn{1}{c}{Solar} & \multicolumn{1}{c}{Traffic} \\
    \midrule
    \multirow{2}{*}{DiffPTS} & CRPS  & \textbf{0.374} & \textbf{0.355} & \textbf{0.329} & \textbf{0.254} & \textbf{0.218 } & \textbf{0.275} & \textbf{0.731} & \textbf{0.173} & \textbf{0.270} \\
          & MSE   & \textbf{0.507} & \textbf{0.452} & \textbf{0.424} & \textbf{0.242} & \textbf{0.180} & \textbf{0.242} & \textbf{2.636} & \textbf{0.179} & \textbf{0.456} \\
    \midrule
    \multirow{2}{*}{w/o NLL $f_\phi$} & CRPS  & 0.553  & 0.943  & 0.540  & 0.813  & 0.544  & 0.944  & 1.409  & 0.494  & 0.540  \\
          & MSE   & 1.125  & 3.178  & 1.101  & 2.527  & 1.050  & 3.231  & 6.493  & 0.781  & 1.431  \\
    \midrule
   \multirow{2}{*}{w/o NLL $g_\psi$} & CRPS  & 0.416  & 0.407  & 0.330  & 0.279  & 0.389  & 0.356  & 0.800  & 0.324  & 0.428  \\
          & MSE   & 0.631  & 0.538  & 0.446  & 0.294  & 0.361  & 0.312  & 2.657  & 0.211  & 0.653  \\
    \midrule
    \multirow{2}{*}{w/o denoise} & CRPS  & 0.437  & 0.369  & 0.387  & 0.288  & 0.245  & 0.346  & 0.888  & 0.190  & 0.277  \\
          & MSE   & 0.631  & 0.469  & 0.498  & 0.290  & 0.201  & 0.343  & 3.271  & 0.208  & 0.626  \\
    \bottomrule
    \end{tabular}%
  \label{tab:ablation_main}%
\end{table}%


\paragraph{Plug-and-Play Experiments}
\label{subsec:plugin_experiment}

As stated in Remark~\ref{remark:model_equivalence}, our proposed ELBO provides a unified objective for recent diffusion-based methods. To investigate this, we conduct plug-and-play enhancement experiments on NsDiff, RDIT, D3U. For D3U, we explicitly introduce the missing variance backbone $g_\psi(\mathbf{X})$ to enable full compatibility with our framework. The results, summarized in Table~\ref{tab:plugin_play_results}, demonstrate consistent improvements across all models and datasets. Notably, the enhanced D3U achieves performance that surpasses the results reported in Table~\ref{tab:main_results}, attributing to D3U’s strong patch-based mean backbone $f_\phi$~\cite{nie2022time}. Specifically, on the ETTm1 dataset, D3U+DiffPTS reduces CRPS from 0.412 to 0.290 and MSE from 0.589 to 0.357. The results highlight the effectiveness of the ELBO on various methods. Full results including TMDM are provided in Appendix~\ref{adpx:subsec:full_plug_play_results}.  We provide an efficiency analysis in Appendix~\ref{apdx:efficiency_analysis} to quantify the  plugin overhead of DiffPTS.


\begin{table}[htbp]
  \centering
  \footnotesize
  \setlength{\tabcolsep}{5.1pt}
    \caption{Plug-and-play results on NsDiff, RDIT, and D3U. \textbf{Bold face} indicates the best performance.}
    \begin{tabular}{c|cc|cc|cc|cc|cc|cc}
    \toprule
    Models & \multicolumn{2}{c}{NsDiff} & \multicolumn{2}{c|}{+DiffPTS} & \multicolumn{2}{c}{RDIT} & \multicolumn{2}{c|}{+DiffPTS} & \multicolumn{2}{c}{D3U} & \multicolumn{2}{c}{+DiffPTS} \\
    Datasets & CRPS  & \multicolumn{1}{c}{MSE} & CRPS  & MSE   & CRPS  & \multicolumn{1}{c}{MSE} & CRPS  & MSE   & CRPS  & \multicolumn{1}{c}{MSE} & CRPS  & MSE \\
    \midrule
    ETTh1 & 0.392  & 0.594  & \textbf{0.364 } & \textbf{0.576 } & 0.410  & 0.570  & \textbf{0.338 } & \textbf{0.458 } & 0.447  & 0.684  & \textbf{0.349 } & \textbf{0.467 } \\
    ETTh2 & 0.358  & 0.514  & \textbf{0.355 } & \textbf{0.495 } & 0.380  & 0.463  & \textbf{0.363 } & \textbf{0.441 } & 0.385  & 0.520  & \textbf{0.331 } & \textbf{0.369 } \\
    ETTm1 & 0.346  & 0.488  & \textbf{0.339 } & \textbf{0.472 } & 0.336  & 0.504  & \textbf{0.320 } & \textbf{0.375 } & 0.412  & 0.589  & \textbf{0.290 } & \textbf{0.357 } \\
    ETTm2 & 0.256  & 0.281  & \textbf{0.254 } & \textbf{0.278 } & 0.258  & 0.292  & \textbf{0.245 } & \textbf{0.216 } & 0.282  & 0.286  & \textbf{0.242 } & \textbf{0.231 } \\
    \bottomrule
    \end{tabular}%
  \label{tab:plugin_play_results}%
\end{table}%

\section{Conclusion}
\label{sec:conclusion}
In this paper, we rethink the ELBO for diffusion-based probabilistic time series forecasting under the location-scale noise model. We show that the full ELBO naturally induces a Gaussian negative log-likelihood objective for adaptive mean and variance estimators, unifying recent methods and revealing that key terms are often overlooked. Based on this, we propose DiffPTS, an end-to-end framework that jointly optimizes all ELBO components. Experiments demonstrate consistent improvements over existing approaches, with plug-in enhancements surpassing state-of-the-art results. 

\textbf{Limitation \& Future work} Our work focuses on the optimization objective rather than network architecture design; we do not investigate what architectural choices are best suited for modeling in the proposed framework. Future work could explore architecture-aware co-design of estimators and the diffusion process, extend the framework for faster inference, or incorporate domain-specific inductive biases for improved calibration and scalability. See Appendix~\ref{apdx:sec:further} for further analyses.

\bibliographystyle{IEEEtran}
\bibliography{nips2027_camera_ready}








\appendix
\clearpage
\section{Derivations}
\label{apdx:sec:deriviation}
\subsection{Proposition \ref{proposition:complete_elbo}}
\label{sec:proof:proposition:complete_elbo}

Without loss of generality, we consider the endpoint model from NsDiff~\cite{ye2025non} with a standard noise schedule, which has the following endpoint distribution and  forward process:

\begin{equation}
    p(\mathbf{Y}_T \mid \mathbf{X}) := \mathcal{N}\big( f_\phi(\mathbf{X}),\, g_\psi(\mathbf{X})) \big.,
\label{eq:pT_endpoint}
\end{equation}

\begin{equation}
\mathbf{Y}_t = \sqrt{\bar{\alpha}_t} \mathbf{Y}_0 + (1 - \sqrt{\bar{\alpha}_t}) f_\phi(\mathbf{X}) + \sqrt{(1 - \bar{\alpha}_t) g_\psi(\mathbf{X})} \, \boldsymbol{\epsilon}_0,
\label{eq:forward_Yt}
\end{equation}
where $\boldsymbol{\epsilon}_0 \sim \mathcal{N}(\mathbf{0}, \mathbf{I})$, $\bar{\alpha}_t = \prod_{i=1}^t (1 - \beta_i)$ with a fixed noise schedule, and $f_\phi$, $g_\psi$ are learnable neural networks with parameters $\phi$ and $\psi$. The ELBO admits the following form~\cite{nakkiran2025step, kingma2013auto}:

\begin{align}
\text{ELBO}_{\theta, \phi, \psi} 
&=\; \underbrace{\mathbb{E}_{q(\mathbf{Y}_1 \mid \mathbf{Y}_0, \mathbf{X})} \left[ \log p_\theta(\mathbf{Y}_0 \mid \mathbf{Y}_1, \mathbf{X}) \right]}_{\mathcal{L}_{\text{recon}} \text{ (reconstruction)}} \nonumber   - \underbrace{D_{\text{KL}}\big( q(\mathbf{Y}_T \mid \mathbf{Y}_0, \mathbf{X}) \,\|\, p(\mathbf{Y}_T \mid \mathbf{X}) \big)}_{\mathcal{L}_{\text{prior}} \text{ (prior matching)}} \\
&- \underbrace{\sum_{t=2}^{T} \mathbb{E}_{q(\mathbf{Y}_t \mid \mathbf{Y}_0, \mathbf{X})} \left[ D_{\text{KL}}\big( q(\mathbf{Y}_{t-1} \mid \mathbf{Y}_t, \mathbf{Y}_0, \mathbf{X}) \,\|\, p_\theta(\mathbf{Y}_{t-1} \mid \mathbf{Y}_t, \mathbf{X}) \big) \right]}_{\mathcal{L}_{\text{denoise}} \text{ (denoising)}}
\label{eq:elbo_standard_apdx}
\end{align}
We first derive the posterior distribution and then analyze these three components individually.

\begin{figure}[htbp]
\centering
\begin{tikzpicture}[
  every node/.style={font=\small},
  lat/.style={circle, draw, minimum size=0.95cm, inner sep=1pt},
  obs/.style={circle, draw, fill=gray!25, minimum size=0.95cm, inner sep=1pt},
  det/.style={rectangle, draw, rounded corners=2pt, minimum height=0.65cm, inner sep=3pt},
  >={Stealth[length=2mm]}]
  \node[lat] (yT)  at (0,0)   {$\mathbf{Y}_T$};
  \node[lat] (yt)  at (2.9,0) {$\mathbf{Y}_t$};
  \node[lat] (ytm) at (5.8,0) {$\mathbf{Y}_{t-1}$};
  \node[obs] (y0)  at (8.7,0) {$\mathbf{Y}_0$};
  \node[obs] (x)   at (4.35,-2.3) {$\mathbf{X}$};
  \node[det] (f)   at (-2.3,-1.2) {$f_\phi(\mathbf{X})$};
  \node[det] (g)   at (-2.3,0.8)  {$g_\psi(\mathbf{X})$};

  \draw[->] (yT) -- node[above]{$\cdots$} (yt);
  \draw[->] (yt) -- node[below=3pt, fill=white, inner sep=1.5pt]{$p_\theta(\mathbf{Y}_{t-1}\!\mid\!\mathbf{Y}_t,\mathbf{X})$} (ytm);
  \draw[->] (ytm) -- node[above]{$\cdots$} (y0);
  \draw[->] (x) -- (yT);
  \draw[->] (x) -- (yt);
  \draw[->] (x) -- (ytm);
  \draw[->] (x) -- (y0);
  \draw[->] (x.west) to[bend left=20] (f.south);
  \draw[->] (x.west) to[bend left=10] (g.south);
  \draw[->] (f) -- (yT);
  \draw[->] (g) -- (yT);
  \draw[->, dashed, gray] (ytm) to[bend right=45] node[above=1pt, fill=white, inner sep=1.5pt]{$q(\mathbf{Y}_t\!\mid\!\mathbf{Y}_{t-1},\mathbf{X})$} (yt);
  \draw[->, dashed, gray] (y0) to[bend right=40] (ytm);
  \draw[->, dashed, gray] (yt) to[bend right=40] (yT);
\end{tikzpicture}
\caption{Probabilistic graphical model of the conditional diffusion process in DiffPTS. Shaded nodes are observed and rectangles denote the deterministic estimators. The historical window $\mathbf{X}$ conditions the endpoint $p(\mathbf{Y}_T\mid\mathbf{X})=\mathcal{N}(f_\phi(\mathbf{X}), g_\psi(\mathbf{X}))$ through $f_\phi, g_\psi$, as well as every reverse transition $p_\theta(\mathbf{Y}_{t-1}\mid\mathbf{Y}_t,\mathbf{X})$ (solid arrows). Dashed arrows denote the forward process $q(\mathbf{Y}_t\mid\mathbf{Y}_{t-1},\mathbf{X})$ in Equation~\eqref{eq:forward_Yt_Nsdiff}. This conditional Markov structure yields the posterior factorization derived below.}
\label{fig:pgm}
\end{figure}
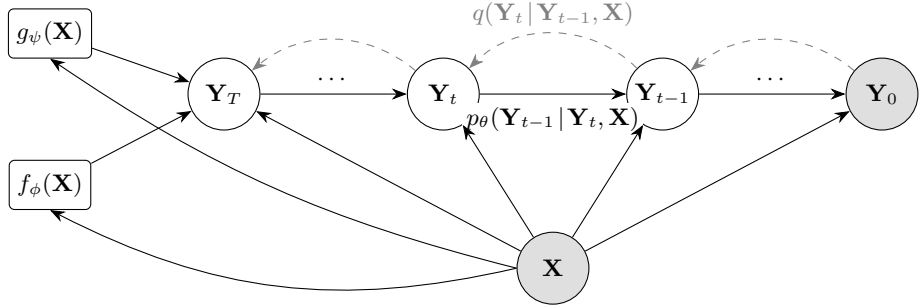

Throughout the derivation, we explicitly assume the one-step Markov property for the forward process conditioned on $\mathbf{X}$, i.e., $q(\mathbf{Y}_{1:T} \mid \mathbf{Y}_0, \mathbf{X}) = \prod_{t=1}^{T} q(\mathbf{Y}_t \mid \mathbf{Y}_{t-1}, \mathbf{X})$, where the dependence on $\mathbf{X}$ enters only through $f_\phi(\mathbf{X})$ and $g_\psi(\mathbf{X})$, together with the corresponding reverse factorization $p_\theta(\mathbf{Y}_{0:T} \mid \mathbf{X}) = p(\mathbf{Y}_T \mid \mathbf{X}) \prod_{t=1}^{T} p_\theta(\mathbf{Y}_{t-1} \mid \mathbf{Y}_t, \mathbf{X})$; hence, given $\mathbf{Y}_t$ and $\mathbf{X}$, $\mathbf{Y}_{t-1}$ is conditionally independent of $\mathbf{Y}_{t+1:T}$ (Figure~\ref{fig:pgm}). The posterior decomposition below follows from the Bayesian rule under these assumptions.

\paragraph{The Posterior Distribution.}
The posterior distribution is derived using the Bayesian rule~\cite{stone2013bayes, hastie2009elements}: 
\begin{align}
q\left(\mathbf{Y}_{t-1} \mid \mathbf{Y}_t, \mathbf{Y}_0, \mathbf{X}\right) \propto q\left(\mathbf{Y}_t \mid \mathbf{Y}_{t-1}, f_\phi(\mathbf{X}), g_\psi(\mathbf{X})\right) q\left(\mathbf{Y}_{t-1} \mid \mathbf{Y}_0, f_\phi(\mathbf{X}), g_\psi(\mathbf{X})\right). 
\end{align}
The mean and variance structure of the equation are independent of each other. ~\cite{ye2025non, li2024tmdm, han2022card} have proved that the posterior $\tilde\mu$ is given in Equation~\eqref{eq:forward_posterior_mean}. ~\cite{ye2025non} consider the ground truth variance $\boldsymbol{\sigma}_\mathbf{Y}$ and prove that the posterior variance $\tilde{\boldsymbol{\sigma}}$ is:
$$
\tilde{\boldsymbol{\sigma}}(\mathbf{X},\boldsymbol{\sigma}_{\mathbf{Y}_0})= \frac{(\beta_t^2 g_\psi(\mathbf{X})+\alpha_t\beta_t \boldsymbol{\sigma}_{\mathbf{Y}_0})(( \bar\beta_{t-1} - \tilde\beta_{t-1})g_\psi(\mathbf{X}) + \tilde\beta_{t-1}\boldsymbol{\sigma}_{\mathbf{Y}_0})}{g_\psi(\mathbf{X})(\alpha_t\bar\beta_{t-1} - \alpha_t\tilde\beta_{t-1} + \beta_t^2) + \boldsymbol{\sigma}_{\mathbf{Y}_0}(\alpha_{t}\tilde\beta_{t-1} + \alpha_t\beta_t)}.
$$
In practice, the ground truth variance $\boldsymbol{\sigma}_{\mathbf{Y}_0}$ is often intractable; hence, we consider a more general case where the variance should be estimated by $g_\psi(\mathbf{X})$, i.e., $g_\psi(\mathbf{X})=\boldsymbol{\sigma}_{\mathbf{Y}_0}$, the posterior variance $\tilde{\boldsymbol{\sigma}}$ can thus be simplified as: $\tilde{\boldsymbol{\sigma}} = \frac{1-\bar{\alpha}_{t-1}}{1-\bar{\alpha}_t} \beta_t g_\psi(\mathbf{X})$. Together, the posterior is:

$$
q(\mathbf{Y}_{t-1} \mid \mathbf{Y}_t, \mathbf{Y}_0, \mathbf{X}) = \mathcal{N}\left( \tilde{\boldsymbol{\mu}}_t(\mathbf{Y}_t, \mathbf{Y}_0, \mathbf{X}),\; \tilde{\boldsymbol{\sigma}}(\mathbf{X}) \right), 
$$
$$
\tilde{\boldsymbol{\mu}}_t:=\frac{\beta_t \sqrt{\bar{\alpha}_{t-1}}}{1-\bar{\alpha}_t} \mathbf{Y}_0+\frac{\left(1-\bar{\alpha}_{t-1}\right) \sqrt{\alpha_t}}{1-\bar{\alpha}_t} \mathbf{Y}_t+\left(1+\frac{\left(\sqrt{\bar{\alpha}_t}-1\right)\left(\sqrt{\alpha_t}+\sqrt{\bar{\alpha}_{t-1}}\right)}{1-\bar{\alpha}_t}\right) f_\phi(\mathbf{X}),
$$
$$
\tilde{\boldsymbol{\sigma}}_t:= \frac{1-\bar{\alpha}_{t-1}}{1-\bar{\alpha}_t} \beta_t g_\psi(\mathbf{X}).
$$

\paragraph{The Reconstruction Term.}

From the definition of the reverse step at $t=1$:
\begin{equation}
    p_\theta(\mathbf{Y}_0 \mid \mathbf{Y}_1, \mathbf{X}) = \mathcal{N}\left( \mu_\theta(\mathbf{Y}_1, \mathbf{X}),\; \beta_1 g_\psi(\mathbf{X})  \right),
    \label{eq:recon_poster_distri}
\end{equation}
where $\mu_\theta$ is the reverse posterior mean defined in Equation \eqref{eq:mu_theta_def}.


We reparameterize $\mu_\theta$ using a noise-predicting network $\epsilon_\theta$. From the forward equation at $t=1$:
\[
\mathbf{Y}_1 = \sqrt{\bar{\alpha}_1} \mathbf{Y}_0 + (1 - \sqrt{\bar{\alpha}_1}) f_\phi(\mathbf{X}) + \sqrt{\beta_1 g_\psi(\mathbf{X})} \, \boldsymbol{\epsilon}_0.
\]

Solving for $\mathbf{Y}_0$:
\begin{equation}
    \mathbf{Y}_0 = \frac{\mathbf{Y}_1 - (1 - \sqrt{\bar{\alpha}_1}) f_\phi(\mathbf{X}) - \sqrt{\beta_1 g_\psi(\mathbf{X})} \, \boldsymbol{\epsilon}_0}{\sqrt{\bar{\alpha}_1}}.
    \label{eq:Y0_in_Y1}
\end{equation}

Thus, we set:
\begin{equation}
    \mu_{\theta}(\mathbf{Y}_1, \mathbf{X}) = \frac{\mathbf{Y}_1 - (1 - \sqrt{\bar{\alpha}_1}) f_\phi(\mathbf{X}) - \sqrt{\beta_1 g_\psi(\mathbf{X})} \, \boldsymbol{\epsilon}_{\theta}(\mathbf{Y}_1, \mathbf{X}, 1)}{\sqrt{\bar{\alpha}_1}}.
    \label{eq:mu_theta_def}
\end{equation}
Plugging  Equation~\eqref{eq:recon_poster_distri} into the Gaussian log-likelihood~\cite{murphy2022probabilistic}:
\begin{equation}
\log p_\theta\left(\mathbf{Y}_0 \mid \mathbf{Y}_1, \mathbf{X}\right)=-\frac{1}{2}\left[\sum_{i=1}^d \log \left(2 \pi \beta_1\left[g_\psi(\mathbf{X})\right]_i\right)+\sum_{i=1}^d \frac{\left(\left[\mathbf{Y}_0-\mu_\theta\right]_i\right)^2}{\beta_1\left[g_\psi(\mathbf{X})\right]_i}\right],
\end{equation}

where $d = N \times D$ is the dimensionality of the data, $N$ is the input length, $D$ is the number of variates in input.


Now compute the squared error by replacing Equation~\eqref{eq:Y0_in_Y1} and ~\eqref{eq:mu_theta_def} into $\mathbf{Y}_0 - \mu_\theta$:
\begin{equation}
    \mathbf{Y}_0 - \mu_\theta = \frac{\sqrt{\beta_1 g_\psi(\mathbf{X})}}{\sqrt{\bar{\alpha}_1}} (\boldsymbol{\epsilon}_0 - \boldsymbol{\epsilon}_{\theta}(\mathbf{Y}_1, \mathbf{X}, 1)).
    \end{equation}
Now plug into the quadratic term:

$$
\frac{\left(\mathbf{Y}_0-\mu_\theta\right)^2}{\beta_1 g_\psi(\mathbf{X})}=\frac{\beta_1 g_\psi(\mathbf{X}) \odot\left(\boldsymbol{\epsilon}_0-\boldsymbol{\epsilon}_\theta\right)^2 / \bar{\alpha}_1}{\beta_1 g_\psi(\mathbf{X})}=\frac{\left(\boldsymbol{\epsilon}_0-\boldsymbol{\epsilon}_\theta\right)^2}{\bar{\alpha}_1}.
$$

Thus, the log-likelihood simplifies as:

$$
\log p_\theta\left(\mathbf{Y}_0 \mid \mathbf{Y}_1, \mathbf{X}\right)=-\frac{1}{2}\left[\sum_{i=1}^d \log \left(2 \pi \beta_1\left[g_\psi(\mathbf{X})\right]_i\right)+\frac{1}{\bar{\alpha}_1}\left\|\boldsymbol{\epsilon}_0-\boldsymbol{\epsilon}_\theta\left(\mathbf{Y}_1, \mathbf{X}, 1\right)\right\|^2\right].
$$

We therefore have the reconstruction loss as:

\begin{align}
    \mathcal{L}_{\text{recon}} &= \frac{1}{2}\mathbb{E}_{q(\mathbf{Y}_1 \mid \mathbf{Y}_0, \mathbf{X})} \left[\sum_{i=1}^d \log \left(2 \pi \beta_1\left[g_\psi(\mathbf{X})\right]_i\right)+\frac{1}{\bar{\alpha}_1}\left\|\boldsymbol{\epsilon}_0-\boldsymbol{\epsilon}_\theta\left(\mathbf{Y}_1, \mathbf{X}, 1\right)\right\|^2  \right]\\
    &= \frac{1}{2} \sum_{i=1}^d\log(2\pi \beta_1 \left[g_\psi(\mathbf{X})\right]_i)+\mathbb{E}_{q(\mathbf{Y}_1 \mid \mathbf{Y}_0, \mathbf{X})} \left[ \frac{1}{2\bar{\alpha}_1} \| \boldsymbol{\epsilon}_0 - \boldsymbol{\epsilon}_{\theta,\phi,\psi}(\mathbf{Y}_1, \mathbf{X}, 1) \|^2 \right] \label{eq:reconstruct_final}\\
    &\propto \frac{1}{2}\sum_{i=1}^d\log \left[g_\psi(\mathbf{X})\right]_i + \mathbb{E}_{q(\mathbf{Y}_1 \mid \mathbf{Y}_0, \mathbf{X})}[\| \boldsymbol{\epsilon}_0 - \boldsymbol{\epsilon}{_{\theta,\phi,\psi}}(\mathbf{Y}_1, \mathbf{X}, 1) \|^2]. \label{eq:reconstruct_propto_final}
\end{align}
\paragraph{The Prior Matching Term.}
The forward distribution at the final time step $t=T$ in Equation~\eqref{eq:forward_Yt} is:

\begin{equation}
q(\mathbf{Y}_T \mid \mathbf{Y}_0, \mathbf{X}) = \mathcal{N}\left( \sqrt{\bar{\alpha}_T} \mathbf{Y}_0 + (1 - \sqrt{\bar{\alpha}_T}) f_\phi(\mathbf{X}),\; \bar{\beta}_T g_\psi(\mathbf{X}) \right),\label{eq:forward_T}
\end{equation}
where $\bar{\beta}_T = 1 - \bar{\alpha}_T$. By substituting the prior defined in Equation~\eqref{eq:pT_endpoint}, the two $d$-dimensional independent Gaussian distributions the KL divergence is:

\begin{align}
& D_{\mathrm{KL}}\left(q\left(\mathbf{Y}_T \mid \mathbf{Y}_0, \mathbf{X}\right) \| p\left(\mathbf{Y}_T \mid \mathbf{X}\right)\right) \\
& =\frac{1}{2} \sum_{i=1}^d\left[\log \frac{\left[g_\psi(\mathbf{X})\right]_i}{\bar{\beta}_T\left[g_\psi(\mathbf{X})\right]_i}+\frac{\bar{\beta}_T\left[g_\psi(\mathbf{X})\right]_i+\left(\left[\sqrt{\bar{\alpha}_T} \mathbf{Y}_0+\left(1-\sqrt{\bar{\alpha}_T}\right) f_\phi(\mathbf{X})-f_\phi(\mathbf{X})\right]_i\right)^2}{\left[g_\psi(\mathbf{X})\right]_i}-1\right] \\
& =\frac{1}{2} \sum_{i=1}^d\left[-\log \bar{\beta}_T+\frac{\bar{\beta}_T\left[g_\psi(\mathbf{X})\right]_i+\left(\sqrt{\bar{\alpha}_T}\left[\mathbf{Y}_0\right]_i-\sqrt{\bar{\alpha}_T}\left[f_\phi(\mathbf{X})\right]_i\right)^2}{\left[g_\psi(\mathbf{X})\right]_i}-1\right] \\
& =\frac{1}{2} \sum_{i=1}^d\left[-\log \left(1-\bar{\alpha}_T\right)-\bar{\alpha}_T+\frac{\bar{\alpha}_T\left(\left[\mathbf{Y}_0\right]_i-\left[f_\phi(\mathbf{X})\right]_i\right)^2}{\left[g_\psi(\mathbf{X})\right]_i}\right] .
\end{align}

Here, the right term can be replaced as a  vector norm, then the prior matching loss term is:

\begin{align}
\mathcal{L}_{\text{prior}} &= D_{\text{KL}}(q \| p) \\
&= \frac{d}{2} \left( -\ln \bar{\beta}_T -\bar{\alpha}_T \right) + \frac{\bar{\alpha}_T}{2}\|\frac{ f_\phi(\mathbf{X}) - \mathbf{Y}_0 }{\sqrt{g_\psi(\mathbf{X})}} \|^2 \label{eq:prior_final} \\
&\propto \frac{1}{2}\|\frac{ f_\phi(\mathbf{X}) - \mathbf{Y}_0 }{\sqrt{g_\psi(
\mathbf{X})}} \|^2. \label{eq:prior_propto_term}
\end{align}



Although $\bar{\alpha}_T$ approaches $0$ as $T \to \infty$, for the finite $T$ used in practice this coefficient is far from negligible, and the term must be optimized. Under our experimental setting with a linear noise schedule ($\beta_{\text{start}} = 10^{-4}$, $\beta_{\text{end}} = 0.02$, and $T = 20$ steps), $\bar{\alpha}_T \approx 0.82$ (for comparison, $\bar{\alpha}_T \approx 0.36$ with $T = 100$), so the prior matching term carries a substantial weight in the ELBO.






\paragraph{The Denoising Matching Terms.}

For each time step $t = 2, \ldots, T$, the denoising matching term in the ELBO is:
\begin{equation}
\mathcal{L}_{\text{denoise}, t>=2} = \sum_{t=2}^T\mathbb{E}_{q(\mathbf{Y}_t \mid \mathbf{Y}_0, \mathbf{X})} \left[ D_{\text{KL}}\big( q(\mathbf{Y}_{t-1} \mid \mathbf{Y}_t, \mathbf{Y}_0, \mathbf{X}) \,\|\, p_\theta(\mathbf{Y}_{t-1} \mid \mathbf{Y}_t, \mathbf{X}) \big) \right] \label{eq:denoise_def}
\end{equation}

Both $q(\mathbf{Y}_{t-1} \mid \mathbf{Y}_t, \mathbf{Y}_0, \mathbf{X})$ and $p_\theta(\mathbf{Y}_{t-1} \mid \mathbf{Y}_t, \mathbf{X})$ are Gaussian with isotropic covariance. The forward posterior is:
\begin{equation}
q(\mathbf{Y}_{t-1} \mid \mathbf{Y}_t, \mathbf{Y}_0, \mathbf{X}) = \mathcal{N}\left( \tilde{\boldsymbol{\mu}}_t(\mathbf{Y}_t, \mathbf{Y}_0, \mathbf{X}),\; \tilde{\beta}_t g_\psi(\mathbf{X}) \right) 
\end{equation}
where $\tilde{\beta}_t = \frac{\bar{\beta}_{t-1}}{\bar{\beta}_t} \beta_t$ and the mean is given in Han et al.~\cite{han2022card}:
\begin{equation}
\tilde{\boldsymbol{\mu}}:=\frac{\beta_t \sqrt{\bar{\alpha}_{t-1}}}{1-\bar{\alpha}_t} \mathbf{Y}_0+\frac{\left(1-\bar{\alpha}_{t-1}\right) \sqrt{\alpha_t}}{1-\bar{\alpha}_t} \mathbf{Y}_t+\left(1+\frac{\left(\sqrt{\bar{\alpha}_t}-1\right)\left(\sqrt{\alpha_t}+\sqrt{\bar{\alpha}_{t-1}}\right)}{1-\bar{\alpha}_t}\right) f_\phi(\mathbf{X})
\end{equation}

Let $\boldsymbol{\mu}_\theta$ be in the form analog to $\tilde{\boldsymbol{\mu}}$ and substitude $\mathbf{Y}_t$ in Equation~\eqref{eq:forward_Yt} into Equation~\eqref{eq:forward_posterior_mean}, we obtain:


\begin{equation}
\label{eq:mean_subtracted}
\begin{aligned}
\boldsymbol{\mu}_\theta - \tilde{\boldsymbol{\mu}} &= \frac{(1 - \alpha_t) \sqrt{\bar{\alpha}_{t-1}}}{\sqrt{\bar\alpha_t}(1-\bar{\alpha}_t)}  \sqrt{(1-\bar\alpha_{t})g_\psi(\mathbf{X})}(\boldsymbol{\epsilon} - \boldsymbol{\epsilon}_\theta) \\
\end{aligned}
\end{equation}

Since $q$ and $p$ share the same variance, the KL divergence between these two Gaussians reduces to:
\begin{equation}
D_{\text{KL}}\big( q(\mathbf{Y}_{t-1} \mid \mathbf{Y}_t, \mathbf{Y}_0, \mathbf{X}) \,\|\, p_\theta(\mathbf{Y}_{t-1} \mid \mathbf{Y}_t, \mathbf{X}) \big) = \frac{1}{2} \sum_{i=1}^d \frac{\left(\tilde{\mu}_{t, i}-\mu_{\theta, i}\right)^2}{\tilde{\beta}_t\left[g_\psi(\mathbf{X})\right]_i} .\label{eq:kl_denoise}
\end{equation}

Replace Equation~\eqref{eq:forward_posterior_mean} into the above formula, the result is equivalent to that of the DDPM~\cite{ye2025non, luo2022understanding, han2022card}:

\begin{align}
\mathcal{L}_{\text{denoise}, t} &= \sum_{t=2}^T \gamma_t \mathbb{E}_{q} \left[ \| \boldsymbol{\epsilon} - \boldsymbol{\epsilon}_\theta(\mathbf{Y}_t, \mathbf{X}, t) \|^2 \right] \label{eq:denoise_final}\\
&\propto \sum_{t=2}^T\mathbb{E}_{q} \left[ \left\| \boldsymbol{\epsilon}_0 - \boldsymbol{\epsilon}_\theta(\mathbf{Y}_t, \mathbf{X}, t) \right\|^2 \right], \label{eq:denoise_propto_final}
\end{align}
where $\gamma_t=\frac{\beta_t}{2 \alpha_t\left(1-\bar{\alpha}_{t-1}\right)}$, when $t\geq 2$.

\paragraph{Combining Three Terms.}

Combining the three loss terms in Equations~\eqref{eq:reconstruct_final}, ~\eqref{eq:prior_final}, and ~\eqref{eq:denoise_final}, we have:

\begin{align}
    -\mathrm{ELBO} &= \frac{1}{2} \sum_{i=1}^d\log(2\pi \beta_1 \left[g_\psi(\mathbf{X})\right]_i)+\mathbb{E}_{q(\mathbf{Y}_1 \mid \mathbf{Y}_0, \mathbf{X})} \left[ \frac{1}{2\bar{\alpha}_1} \| \boldsymbol{\epsilon}_0 - \boldsymbol{\epsilon}_{\theta,\phi,\psi}(\mathbf{Y}_1, \mathbf{X}, 1) \|^2 \right] \\
    &+ \frac{d}{2} \left( -\ln \bar{\beta}_T -\bar{\alpha}_T \right) + \frac{\bar{\alpha}_T}{2}\|\frac{ f_\phi(\mathbf{X}) - \mathbf{Y}_0 }{\sqrt{g_\psi(\mathbf{X})}} \|^2\\
    &+  \sum_{t=2}^T \gamma_t \mathbb{E}_{q} \left[ \| \boldsymbol{\epsilon}_0 - \boldsymbol{\epsilon}_\theta(\mathbf{Y}_t, \mathbf{X}, t) \|^2 \right] \\
    &=C + \sum_{t=1}^{T} \gamma_t \mathbb{E}_{\boldsymbol{\epsilon}_0} \left[ \| \boldsymbol{\epsilon}_0 - \boldsymbol{\epsilon}_\theta(\mathbf{Y}_t, \mathbf{X}, t) \|^2 \right] + \frac{\bar{\alpha}_T}{2}\|\frac{ f_\phi(\mathbf{X}) - \mathbf{Y}_0 }{\sqrt{g_\psi(\mathbf{X})}} \|^2 + \frac{1}{2} \sum_{i=1}^d\log \left[g_\psi(\mathbf{X})\right]_i
\end{align}
where $C$ is some constants, $\gamma_t=\frac{\beta_t}{2 \alpha_t\left(1-\bar{\alpha}_{t-1}\right)} \text { for } t>1, \text { and } \gamma_1=\frac{1}{2 \alpha_1}$.  This completes the proof.

\subsection{Remark \ref{remark:model_equivalence}}
\label{sec:proof:model_equivalence}

The ELBO formulation is determined by the definition of the forward process and the denoise schedule which decides the reverse posterior~\cite{lee2021priorgrad}. We prove the general equivalence  under standard noise schedule by proving that these methods share the same forward process in this condition.  First, we recap on the forward process of DiffPTS:

\begin{equation}
\mathbf{Y}_t = \sqrt{\bar{\alpha}_t} \mathbf{Y}_0 + (1 - \sqrt{\bar{\alpha}_t}) f_\phi(\mathbf{X}) + \sqrt{(1 - \bar{\alpha}_t) g_\psi(\mathbf{X})} \, \boldsymbol{\epsilon}.
\label{eq:apdx_forward}
\end{equation}

\textbf{Endpoint-based.} NsDiff~\cite{ye2025non} additionally introduces an uncertainty-aware noise schedule based on the LSNM assumption, which defines a forward noise schedule as:  
\begin{equation}
\mathbf{Y}_t =  \sqrt{\bar\alpha_t}\mathbf{Y}_0 + (1 - \sqrt{\bar\alpha_t})f_\phi(\mathbf{X}) + \sqrt{ ( \bar\beta_t - \tilde\beta_t)g_\psi(\mathbf{X}) + \tilde\beta_t\boldsymbol{\sigma}_{\mathbf{Y}_0}}\boldsymbol{\epsilon}.
\label{eq:apdx_forward_nsdiff}
\end{equation}

However, it requires prior knowledge of the ground truth variance ${\sigma}_{\mathbf{Y}_0}$, which is often intractable in regular time series~\cite{hyndman2018forecasting, liu2021forecast}. In a more general case where the denoise schedule follows the standard DDPM (i.e., $g_\psi = {\sigma}_{\mathbf{Y}_0}$), Equation~\eqref{eq:apdx_forward_nsdiff} naturally reduces to the forward definition of DiffPTS, which we refer to as the standard-schedule instantiation of NsDiff. The uncertainty-aware schedule is complementary to our formulation. It relies on label information for the variance, namely the future sliding-window variance used as a supervised target, whereas our objective trains the variance estimator without labels through the ELBO, and the standard-schedule NsDiff+DiffPTS already improves NsDiff (Table~\ref{tab:plugin_play_results}). Combining the label-free objective with an uncertainty-aware schedule is left for future work.

\textbf{Residual-based.} We  prove that the ELBO of RDIT and DiffPTS's diffusion paradigms are equivalent under a simple variable transformation. 
Since the variance $g(\mathbf{X}) \neq 0$, Equation~\eqref{eq:apdx_forward} leads to:
\begin{equation}
    \frac{\mathbf{Y}_t-f_\phi(\mathbf{X})}{\sqrt{g_\psi(\mathbf{X})}}=\sqrt{\bar{\alpha}_t} \cdot \frac{\mathbf{Y}_0-f_\phi(\mathbf{X})}{\sqrt{g_\psi(\mathbf{X})}}+\sqrt{1-\bar{\alpha}_t} \boldsymbol{\epsilon}.
\end{equation}
 Define the transformed variable $\mathbf{Z}_t$:
\begin{equation}
\mathbf{Z}_t=\frac{\mathbf{Y}_t-f_\phi(\mathbf{X})}{\sqrt{g_\psi(\mathbf{X})}}.
\label{eq:trainformation_of_variatble_z}
\end{equation}

the forward process and the prior of the transformed process are as follows:

\begin{equation}
\mathbf{Z}_t=\sqrt{\bar{\alpha}_t} \mathbf{Z}_0+\sqrt{1-\bar{\alpha}_t} \boldsymbol{\epsilon}, \quad \boldsymbol{\epsilon} \sim \mathcal{N}(\mathbf{0}, \mathbf{I})
\label{eq:z_standard_denoise}
\end{equation}
\begin{equation}
p\left(\mathbf{Z}_T\right)=\mathcal{N}(\mathbf{0}, \mathbf{I}).
\end{equation}


which is exactly the forward process of residual-based methods~\cite{li2025diffusion,lai2025rdit}. Similarly, $p(\mathbf{Z}_T) = \mathcal{N}(\mathbf{0}, \mathbf{I})$ naturally leads to Equation~\eqref{eq:pT_endpoint} by applying the Gaussian expectation and variance rules. As the ELBO is fully determined by the prior distribution and the forward process, both formulations yield identical objectives. When $g_\psi(\mathbf{X}) = 1$, the diffusion paradigm of RDIT reduces to that of D3U, and the diffusion paradigm of NsDiff reduces to that of TMDM. When further $f_\phi(\mathbf{X}) = \mathbf{0}$, both paradigms reduce to the standard DDPM~\cite{rasul2021autoregressive, shen2023non}.

This completes the proof.

\subsection{Proposition~\ref{prop:loss_function_DiffPTS}}
\label{apdx:proof:prop:loss_function_DiffPTS}

\paragraph{Optimization Objective.}

Summing the reconstruction term \eqref{eq:reconstruct_propto_final}, the prior matching term \eqref{eq:prior_propto_term} and the denoising terms \eqref{eq:denoise_propto_final} for $t=2,\ldots,T$:
\begin{equation}
\begin{aligned}
-\text{ELBO} \propto &\; \frac{1}{2}\underbrace{ \sum_{i=1}^d\log \left[g_\psi(\mathbf{X})\right]_i + \mathbb{E}_{q(\mathbf{Y}_1 \mid \mathbf{Y}_0, \mathbf{X})}[\| \boldsymbol{\epsilon}_0 - \boldsymbol{\epsilon}{_{\theta,\phi,\psi}}(\mathbf{Y}_1, \mathbf{X}, 1) \|^2]}_{\mathcal{L}_{\text{recon}}} \\
& + \underbrace{\frac{1}{2}\|\frac{ f_\phi(\mathbf{X}) - \mathbf{Y}_0 }{\sqrt{g_\psi(
\mathbf{X})}} \|^2}_{\mathcal{L}_{\text{prior matching}}}  + \underbrace{\sum_{t=2}^{T}   \mathbb{E}_{q} \left[ \left\| \boldsymbol{\epsilon}_0 - \boldsymbol{\epsilon}_\theta(\mathbf{Y}_t, \mathbf{X}, t) \right\|^2 \right]}_{\mathcal{L}_{\text{denoise}, t \geq 2} }.
\end{aligned} \label{eq:elbo_sum}
\end{equation}

\begin{equation}
\begin{aligned}
\log p(\mathbf{Y}|\mathbf{X}) \leq \text{ELBO} = (\mathcal{L}_{recon} + \mathcal{L}_{denoise}+\mathcal{L}_{prior})\end{aligned} \label{eq:elbo_bound}
\end{equation}

Combine the denoising terms for all $t = 1, \ldots, T$:
\begin{align}
\mathcal{L}_{\text{denoise}} &=  \sum_{t=1}^{T} \mathbb{E}_{q} \left[ \| \boldsymbol{\epsilon}_0 - \boldsymbol{\epsilon}_\theta(\mathbf{Y}_t, \mathbf{X}, t) \|^2 \right] \label{eq:mse_total} \\
&= \mathbb{E}_{(\mathbf{X}, \mathbf{Y_0})\sim q(\mathbf{X}, \mathbf{Y_0}), t \sim U\{1, T\}} \left[ \| \boldsymbol{\epsilon}_0 - \boldsymbol{\epsilon}_\theta(\mathbf{Y}_t, \mathbf{X}, t) \|^2 \right], \label{eq:mse_total_univorm}
\end{align}
where Equation ~\eqref{eq:mse_total_univorm} is equivalent to Equation ~\eqref{eq:ddpm_denoise_loss}.




Therefore, by combining these three terms in Equations~\eqref{eq:reconstruct_propto_final}, ~\eqref{eq:prior_propto_term}, and ~\eqref{eq:denoise_propto_final}, maximizing the ELBO is equivalent to minimizing the following loss function:

\begin{equation}
\mathcal{L}_{\text{DiffPTS}}(\theta, \phi, \psi) = \frac{1}{2}\sum_{i=1}^d\log \left[g_\psi(\mathbf{X})\right]_i +  \frac{1}{2}\|\frac{ f_\phi(\mathbf{X}) - \mathbf{Y}_0 }{\sqrt{g_\psi(
\mathbf{X})}} \|^2 +  \mathbb{E}_{ t,q} \left[ \| \boldsymbol{\epsilon}_0 - \boldsymbol{\epsilon}_\theta(\mathbf{Y}_t, \mathbf{X}, t) \|^2 \right].   \label{eq:loss_total}
\end{equation}

The expectations are approximated by Monte Carlo sampling~\cite{hung2024review, shapiro2003monte}: for each mini-batch of $(\mathbf{X}, \mathbf{Y}_0)$ pairs, we sample $t \sim \text{Uniform}\{1,\ldots,T\}$ and $\boldsymbol{\epsilon}_0 \sim \mathcal{N}(0, \mathbf{I})$, then compute $\mathbf{Y}_t$ via the forward process. The first term trains the denoising network $\boldsymbol{\epsilon}_\theta$.

Next, We show that the first and second terms together form a negative likelihood objective that trains the mean predictor $f_\phi$ and the variance predictor $g_\psi$ on a gaussian assumption on $p(\mathbf{Y}|\mathbf{X})$.

\paragraph{Conditional Gaussian Negative Log-Likelihood.}
\label{app:gaussian_nll}
Consider a conditional generative model where the target variable $\mathbf{Y}_0$ is modeled as a Gaussian distribution conditioned on covariates $\mathbf{X}$:
\[
p(\mathbf{Y}_0 \mid \mathbf{X}; \phi, \psi) = \mathcal{N}\big( \mathbf{Y}_0 \,\big|\, f_\phi(\mathbf{X}),\; \operatorname{diag}(g_\psi(\mathbf{X})) \big),
\]
where $f_\phi(\mathbf{X})$ is the mean function parameterized by $\phi$, and $g_\psi(\mathbf{X}) \in \mathbb{R}^d_{>0}$ defines the diagonal covariance matrix via $\Sigma(\mathbf{X}) = \operatorname{diag}\big(g_\psi(\mathbf{X})\big)$.

The probability density function of a multivariate Gaussian $\mathcal{N}(\boldsymbol{\mu}, \Sigma)$ is:
\[
p(\mathbf{y}) = \frac{1}{(2\pi)^{d/2} |\Sigma|^{1/2}} 
\exp\!\left( -\frac{1}{2} (\mathbf{y} - \boldsymbol{\mu})^\top \Sigma^{-1} (\mathbf{y} - \boldsymbol{\mu}) \right).
\]

Substituting $\boldsymbol{\mu} = f_\phi(\mathbf{X})$ and $\Sigma = \operatorname{diag}(g_\psi(\mathbf{X}))$, we compute the log-likelihood:
\begin{align}
\log p(\mathbf{Y}_0 \mid \mathbf{X})
&= -\frac{d}{2} \log(2\pi) - \frac{1}{2} \log \det \Sigma(\mathbf{X}) 
   - \frac{1}{2} (\mathbf{Y}_0 - f_\phi(\mathbf{X}))^\top \Sigma(\mathbf{X})^{-1} (\mathbf{Y}_0 - f_\phi(\mathbf{X})) \nonumber \\
&= -\frac{d}{2} \log(2\pi) - \frac{1}{2} \sum_{i=1}^d \log [g_\psi(\mathbf{X})]_i 
   - \frac{1}{2} \sum_{i=1}^d \frac{\big( [\mathbf{Y}_0]_i - [f_\phi(\mathbf{X})]_i \big)^2}{[g_\psi(\mathbf{X})]_i} \nonumber \\
&= -\frac{d}{2} \log(2\pi) - \frac{1}{2} \sum_{i=1}^d \log [g_\psi(\mathbf{X})]_i 
   - \frac{1}{2} \left\| \frac{\mathbf{Y}_0 - f_\phi(\mathbf{X})}{\sqrt{g_\psi(\mathbf{X})}} \right\|^2,
\end{align}
where the division and square root are applied element-wise, and $\|\cdot\|$ denotes the Euclidean norm.

Therefore, the negative log-likelihood (NLL), up to an additive constant independent of the parameters $(\phi, \psi)$, is:
\begin{equation}
-\log p(\mathbf{Y}_0 \mid \mathbf{X}) 
= \frac{1}{2} \left\| \frac{f_\phi(\mathbf{X}) - \mathbf{Y}_0}{\sqrt{g_\psi(\mathbf{X})}} \right\|^2 
+ \frac{1}{2} \sum_{i=1}^d \log [g_\psi(\mathbf{X})]_i 
+ \underbrace{\frac{d}{2} \log(2\pi)}_{\text{constant}}.
\end{equation}

Consequently, minimizing the NLL with respect to $(\phi, \psi)$ is equivalent to minimizing:
\[
\frac{1}{2} \left\| \frac{f_\phi(\mathbf{X}) - \mathbf{Y}_0}{\sqrt{g_\psi(\mathbf{X})}} \right\|^2 
+ \frac{1}{2} \sum_{i=1}^d \log [g_\psi(\mathbf{X})]_i,
\]
which corresponds to the last two terms in Proposition~\ref{prop:loss_function_DiffPTS}.

This completes the proof.

\subsection{Convergence Rate}
\label{apdx:sec:convergence}
Lee et al.~\cite{lee2021priorgrad} demonstrates that the Hessian matrix of a data-dependent Gaussian prior $\mathcal{N}(\mu, \Sigma)$ has a condition number of at least 1. This result naturally extends to our setting. Specifically, under mild constraints and assuming $\epsilon_\theta$ is a linear function, Lee et al.~\cite{lee2021priorgrad} proves that:
$$
 \min_\theta L(\boldsymbol{\mu}, \boldsymbol{\Sigma}, \mathbf{Y}_0; \theta) \leq \min_\theta L(\mathbf{0}, \mathbf{I}, \mathbf{Y}_0; \theta). 
$$
This inequality indicates that a simple model suffices to represent the mean of $q(\mathbf{Y}_{t-1}|\mathbf{Y}_t)$ under a data-dependent prior, whereas achieving comparable precision with an isotropic covariance prior ($\mathbf{0}, \mathbf{I}$) requires a more complex model. Consequently, optimization under the data-dependent prior achieves the optimal condition number of 1 for the Hessian matrix of the loss function, yielding the fastest possible convergence rate.

\section{Experimental Details}

\subsection{Datasets Details}
\label{apdx:subsec:dataset_details}
Following~\cite{ye2025non, li2024tmdm}, we use the following datasets widely used in time series forecasting~\cite{zeng2023transformers, wu2022timesnet,  zhou2021informer}.

\textbf{ETT}\footnote{\url{https://github.com/zhouhaoyi/ETDataset}.} (Electricity Transformer Temperature): The electricity transformer datasets contain power load information and oil temperature as the target variable. \{ETTh1, ETTh2\} are 1-hour-level datasets, and \{ETTm1, ETTm2\} are 15-minute-level datasets. Each data point consists of seven features: one ``oil temperature'' (OT) target and six power load features.

\textbf{ECL}\footnote{\url{ https://github.com/laiguokun/multivariate-time-series-data}.} (Electricity Consuming Load): The electricity consumption dataset contains hourly electricity consumption (in Kwh) of 321 clients in two years. 

\textbf{EXG} (Exchange-Rate): The exchange rate dataset contains daily exchange rates of eight foreign countries including Australia, British, Canada, Switzerland, China, Japan, New Zealand, and Singapore ranging from 1990 to 2016. 

\textbf{Traffic}\footnote{\url{ http://pems.dot.ca.gov}.\label{fn:PEMS}} (Traffic): The data contain 48 months
(2015-2016) of hourly road occupancy rates (between 0 and 1) measured by
862 sensors on freeways in the San Francisco Bay area.

\textbf{Solar}\footnote{\url{https://www.nrel.gov/grid/solar-power-data.html}.} (Solar Energy): The solar power production dataset records the hourly energy output from 137 photovoltaic (PV) plants in Alabama over the year 2006.

\textbf{ILI}\footnote{\url{https://gis.cdc.gov/grasp/fluview/fluportaldashboard.html}.} (Influenza-like Illness): The ILI dataset contains weekly percentages of patients reported with influenza-like illness with multiple flu seasons containing 7 features.

\subsection{Detail Implementation}
\label{apdx:subsec:detail_implementation}
We follow previous works~\cite{ye2025non, li2024tmdm} to use the Non-stationary Transformer~\cite{liu2022non} and a three-layer MLP as implementations of $f_\phi$ and $g_\psi$, respectively. For $f_\phi$, we strictly follow the original Non-stationary Transformer architecture. The variance estimator $g_\psi$ is parameterized by a three-layer MLP with hidden dimension $h=512$:
$$
\mathbf{H}^{(1)} = \text{ReLU}\left(\mathbf{W}^{(1)} \mathbf{X}^\top + \mathbf{b}^{(1)}\right), \quad
\mathbf{H}^{(2)} = \text{ReLU}\left(\mathbf{W}^{(2)} \mathbf{H}^{(1)} + \mathbf{b}^{(2)}\right)\\
$$
$$
g_\psi(\mathbf{X}) = \text{Softplus}\left(\mathbf{W}^{(3)} \mathbf{H}^{(2)} + \mathbf{b}^{(3)}\right) \\
$$
where $\mathbf{W}^{(i)}, \mathbf{b}^{(i)},\mathbf{H}^{(i)}$ are learnable parameters and output matrix of the $i$-th layer. Softplus activation $\text{softplus}(x) = \log(1 + e^x)$~\cite{zheng2015improving} is used to ensure positive outputs for variance estimation while maintaining smooth gradients, which is crucial for stable training of the variance model~\cite{shridhar2018uncertainty}. A variance floor of $10^{-8}$ is further applied for numerical stability (Appendix~\ref{apdx:subsec:stability}).
\subsection{Metrics}
\label{apdx:metrics}
We adopt the following standard metrics to evaluate model performance:


\textbf{Mean Absolute Error (MAE)} measures the average magnitude of absolute prediction errors:
    \[
        \text{MAE} = \frac{1}{MD} \sum_{i=1}^{M}\sum_{j=1}^{D} |\mathbf{Y}_{i,j} - \hat{\mathbf{Y}}_{i,j}|,
    \]
    where $\mathbf{Y}_{i,j}$ and $\hat{\mathbf{Y}}_{i,j}$ denote the ground-truth and predicted values at the $i$-th future step of the $j$-th variate, $M$ is the prediction length, and $D$ is the number of variates. Both MAE and MSE are averaged over all test samples.

\textbf{Mean Squared Error (MSE)} quantifies the average squared deviation between predictions and ground truth, penalizing larger errors more heavily:
    \[
        \text{MSE} = \frac{1}{MD} \sum_{i=1}^{M}\sum_{j=1}^{D}  (\mathbf{Y}_{i,j} - \hat{\mathbf{Y}}_{i,j})^2.
    \]


\textbf{CRPS.} The continuous ranked probability score (CRPS)~\cite{matheson1976scoring} quantifies the discrepancy between a predictive cumulative distribution function (CDF) $F$ and an observed outcome $x$ via
\begin{equation}
\operatorname{CRPS}(F, x)=\int_{\mathbb{R}}\bigl(F(z)-\mathbb{I}\{x \leq z\}\bigr)^2 \,\mathrm{d}z,
\end{equation}
where $\mathbb{I}\{\cdot\}$ denotes the indicator function. In practice, when $F$ is approximated by its empirical counterpart $\hat{F}(z) = \frac{1}{S} \sum_{s=1}^S \mathbb{I}\{ x^{0,s} \leq z \}$constructed from $S$ samples $x^{0,s} \sim F$, the CRPS can be estimated directly from the generated trajectories. In our experiments, we use $S = 100$ samples to approximate $F$.

\textbf{QICE.} The Quantile Interval Calibration Error (QICE)~\cite{han2022card} quantifies the mismatch between the empirical coverage of predicted quantile intervals (QIs) and their nominal target coverage. Specifically, the predictive distribution is partitioned into $I$ non-overlapping, equal-probability quantile intervals, each assigned a nominal coverage of $1/I$. QICE then measures the average absolute deviation between the observed frequency of ground-truth values falling within each interval and the ideal coverage $1/I$.
Formally, let $\hat{y}_n^{\text{low}_m}$ and $\hat{y}_n^{\text{high}_m}$ denote the lower and upper bounds of the $m$-th quantile interval for the $n$-th sample. The empirical coverage of the $m$-th interval is given by
\[
r_m = \frac{1}{N} \sum_{n=1}^N \mathbb{I}\left\{ \hat{y}_n^{\text{low}_m} \leq y_n \leq \hat{y}_n^{\text{high}_m} \right\},
\]
where $\mathbb{I}\{\cdot\}$ is the indicator function. QICE is defined as
\begin{equation}
    \operatorname{QICE} := \frac{1}{I} \sum_{m=1}^I \left| r_m - \frac{1}{I} \right|.
\end{equation}
Under perfect calibration (i.e., when the predicted distribution coincides with the true data-generating distribution), we expect $r_m \approx 1/I$ for all $m$, resulting in $\operatorname{QICE} \to 0$. Following Li et al.~\cite{li2024tmdm}, we use $I = 10$ equal-sized decile-based intervals to compute QICE.



\section{Additional Showcases}
\label{apdx:sec:other_showcases}

\begin{figure}[ht]
\begin{center}
\centerline{\includegraphics[width=1\linewidth]{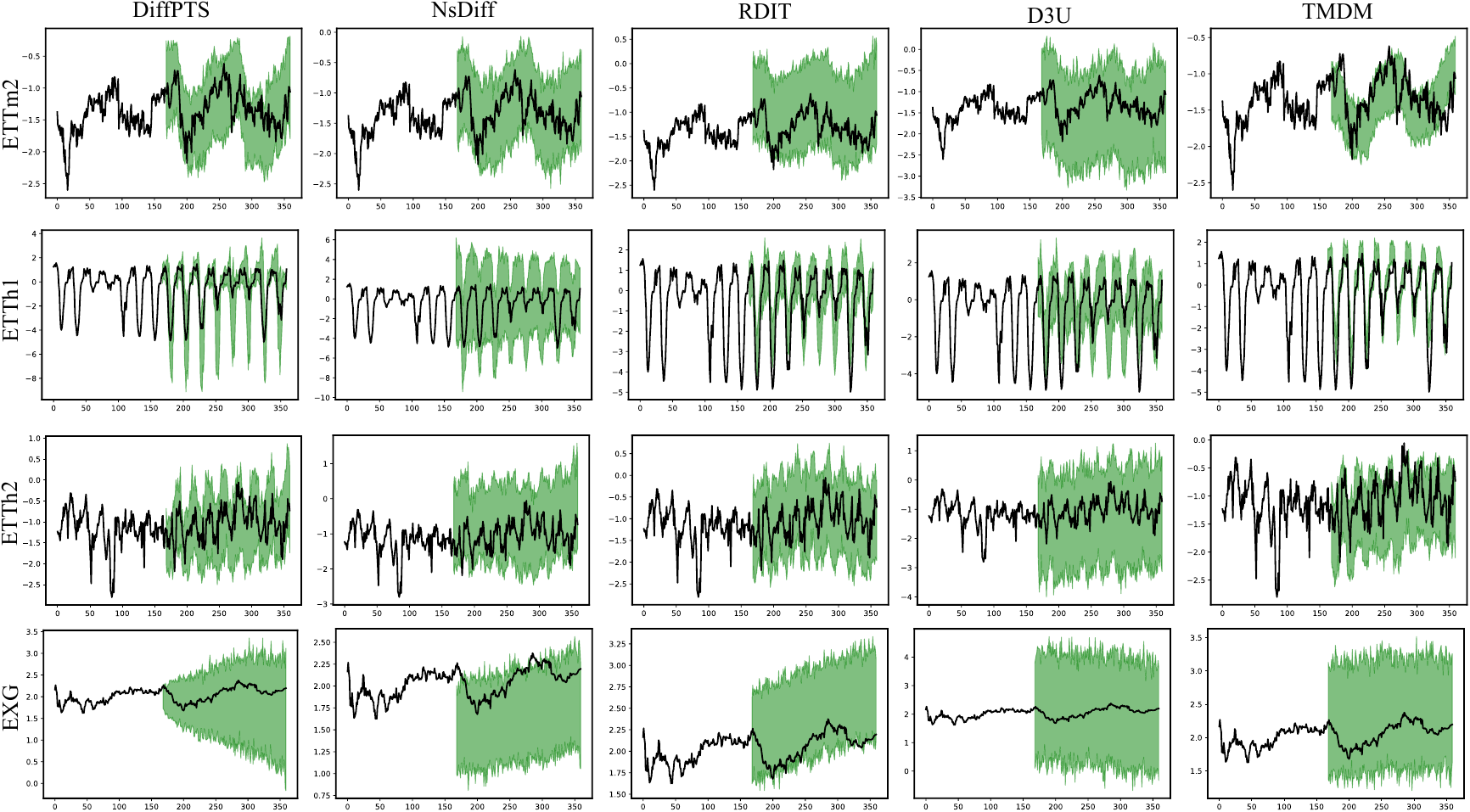}}
\caption{The 96\%  prediction intervals of ETTm2, ETTh1, ETTh2 and EXG samples.}
\label{fig:other_showcase}
\end{center}
\vskip -0.3in
\end{figure}

Figure~\ref{fig:other_showcase} demonstrates DiffPTS's capability in capturing dynamic trend uncertainties across diverse time series. The 96\% prediction intervals (green shaded regions) for four datasets ETTm2, ETTm1, ETTh2, and EXG, with black lines representing ground truth. Specifically, on the ETT datasets, intervals adapt to seasonal patterns, expanding during high-variability periods and contracting during stable trends, demonstrating context-aware uncertainty quantification. Moreover, compared to other methods, DiffPTS explicitly captures bidirectional trend uncertainties in EXG (financial data often exhibits random walk~\cite{lawler2010random}), with intervals widening precisely during sharp upward/downward transitions, reflecting heightened uncertainty during directional changes while maintaining tight coverage.

\section{Full Results}
\subsection{Main Experiments}
\label{subsec:exp_rst:main_exp}

In the appendix, we present comprehensive experimental results across nine benchmark datasets for both probabilistic and point forecasting tasks (Tables~\ref{tab:apdx:exp_rst:full_probabilistic} and~\ref{tab:apdx:exp_rst:full_point}). Our proposed DiffPTS framework demonstrates consistent state-of-the-art performance. NsDiff demonstrates competitive QICE values on specific datasets like ETTh2 and ILI. This can be attributed to NsDiff's uncertainty-aware noise scheduling mechanism, which incorporates variance information into the diffusion process. When the underlying time series exhibits predictable variance patterns within the sliding window, this approximation becomes reasonably accurate, leading to improved uncertainty calibration reflected in the QICE metric. However, this advantage is dataset-dependent and limited to scenarios where the sliding-window assumption holds well. In contrast, DiffPTS achieves more consistent performance across diverse datasets by learning uncertainty patterns directly from data without relying on potentially unstable variance approximations, ultimately delivering better overall probabilistic forecasting capabilities despite occasional dataset-specific advantages of NsDiff's heuristic approach.

\begin{table}[htbp]
  \centering
  \setstretch{0.9}
  \footnotesize
  \caption{Probabilistic forecasting performance (CRPS, QICE) across nine datasets. \textbf{Bold face} indicates the best result, and \underline{underline} indicates the second-best result.}
    \setlength{\tabcolsep}{5.4pt}
    \begin{tabular}{ccccccccccc}
    \toprule
    Models & Datasets & ETTh1 & ETTh2 & ETTm1 & ETTm2 & ECL   & EXG   & ILI   & Solar & Traffic \\
    \midrule
    TimeGrad & CRPS  & 0.606  & 1.212  & 0.647  & 0.775  & 0.397  & 0.826  & 1.140  & 0.293  & 0.407  \\
    (2021) & QICE  & 6.731  & 9.488  & 6.693  & 6.962  & 7.118  & 9.464  & 6.519  & 7.378  & 4.581  \\
    \midrule
    CSDI  & CRPS  & 0.492  & 0.647  & 0.524  & 0.817  & 0.577  & 0.855  & 1.244  & 0.432  & 1.418  \\
    (2022) & QICE  & 3.107  & 5.331  & 2.828  & 8.106  & 7.506  & 7.864  & 7.693  & 9.957  & 13.613  \\
    \midrule
    TimeDiff & CRPS  & 0.465  & 0.471  & 0.464  & 0.316  & 0.750  & 0.433  & 1.153  & 0.700  & 0.771  \\
    (2023) & QICE  & 14.931  & 14.813  & 14.795  & 13.385  & 15.466  & 14.556  & 14.942  & 14.914  & 15.439  \\
    \midrule
    DiffusionTS & CRPS  & 0.603  & 1.168  & 0.574  & 1.035  & 0.633  & 1.251  & 1.612  & 0.470  & 0.668  \\
    (2024) & QICE  & 6.423  & 9.577  & 5.605  & 9.959  & 8.205  & 10.411  & 10.090  & 6.627  & 5.958  \\
    \midrule
    TMDM  & CRPS  & 0.452  & 0.383  & 0.375  & 0.289  & 0.461  & 0.336  & 0.967  & 0.350  & 0.557  \\
    (2024) & QICE  & 2.821  & 4.471  & 2.567  & 2.610  & 10.562  & 6.393  & 6.217  & 9.342  & 10.676  \\
    \midrule
    D3U   & CRPS  & 0.447  & 0.385  & 0.412  & 0.282  & 0.439  & 0.354  & 2.097  & 0.488  & 0.507  \\
    (2025) & QICE  & 4.783  & 6.248  & 5.375  & 7.710  & 6.855  & 8.429  & 10.595  & 7.673  & 6.204  \\
    \midrule
    RDIT  & CRPS  & 0.410  & 0.380  & \underline{0.336}  & 0.258  & 0.362  & 0.392  & 0.845  & \underline{0.225}  & 0.495  \\
    (2025) & QICE  & 4.772  & 4.960  & 6.397  & 6.435  & 9.894  & 9.727  & 8.375  & 7.791  & 6.722  \\
    \midrule
    NsDiff & CRPS  & \underline{0.392}  & \underline{0.358}  & 0.346  & \underline{0.256}  & \underline{0.290}  & \underline{0.324}  & \underline{0.806}  & 0.300  & \underline{0.378}  \\
    (2025) & QICE  & \underline{1.470}  & \textbf{2.074 } & \underline{2.041}  & \underline{2.030}  & \underline{6.685}  & \underline{5.930}  & \textbf{5.598} & \underline{6.820}  & \underline{3.601}  \\
    \midrule
    DiffPTS & CRPS  & \textbf{0.374 } & \textbf{0.355 } & \textbf{0.329 } & \textbf{0.254 } & \textbf{0.218 } & \textbf{0.275 } & \textbf{0.731 } & \textbf{0.173 } & \textbf{0.270 } \\
    (ours) & QICE  & \textbf{1.460 } & \underline{2.699}  & \textbf{1.855 } & \textbf{2.020 } & \textbf{1.140 } & \textbf{3.957 } & \underline{5.905}  & \textbf{2.671 } & \textbf{2.842 } \\
    \bottomrule
    \end{tabular}%
  \label{tab:apdx:exp_rst:full_probabilistic}%
\end{table}%

\begin{table}[htbp]
  \centering
  \setstretch{0.9}
  \footnotesize
  \caption{Point forecasting performance (MSE, MAE) across nine datasets. \textbf{Bold face} indicates the best result, and \underline{underline} indicates the second-best result.}
    \begin{tabular}{ccccccccccc}
    \toprule
    Models & Datasets & ETTh1 & ETTh2 & ETTm1 & ETTm2 & ECL   & EXG   & ILI   & Solar & Traffic \\
    \midrule
    TimeGrad & MAE   & 0.813  & 1.496  & 0.831  & 0.967  & 0.504  & 1.058  & 1.414  & 0.446  & 0.535  \\
    (2021) & MSE   & 1.062  & 3.462  & 1.218  & 1.690  & 0.505  & 1.567  & 4.197  & 0.475  & 0.983  \\
    \midrule
    CSDI  & MAE   & 0.708  & 0.900  & 0.752  & 1.069  & 0.822  & 1.081  & 1.481  & 0.675  & 0.925  \\
    (2022) & MSE   & 0.949  & 1.226  & 1.002  & 1.723  & 1.007  & 1.701  & 4.515  & 0.763  & 1.731  \\
    \midrule
    TimeDiff & MAE   & \underline{0.479}  & 0.485  & 0.477  & \underline{0.333}  & 0.764  & 0.446  & 1.169  & 0.713  & 0.784  \\
    (2023) & MSE   & \underline{0.517}  & \underline{0.456}  & 0.537  & \underline{0.268}  & 0.879  & 0.402  & 3.958  & 0.821  & 1.350  \\
    \midrule
    DiffusionTS & MAE   & 0.774  & 1.411  & 0.744  & 1.232  & 0.856  & 1.564  & 1.788  & 0.740  & 0.815  \\
    (2024) & MSE   & 1.089  & 3.273  & 1.030  & 2.372  & 1.072  & 3.628  & 6.053  & 0.749  & 1.473  \\
    \midrule
    TMDM  & MAE   & 0.607  & 0.490  & \underline{0.455}  & 0.395  & 0.359  & 0.430  & 1.175  & 0.316  & 0.425  \\
    (2024) & MSE   & 0.696  & 0.512  & 0.494  & 0.315  & 0.257  & 0.334  & 3.636  & 0.250  & 0.679  \\
    \midrule
    D3U   & MAE   & 0.551  & 0.484  & 0.501  & 0.389  & 0.502  & \underline{0.372}  & 2.631  & 0.613  & 0.605  \\
    (2025) & MSE   & 0.684  & 0.520  & 0.589  & 0.286  & 0.440  & 0.269  & 9.951  & 0.644  & 0.907  \\
    \midrule
    RDIT  & MAE   & 0.535  & \underline{0.481}  & 0.461  & 0.364  & 0.308  & 0.373  & 0.986  & \underline{0.239}  & 0.571  \\
    (2025) & MSE   & 0.570  & 0.463  & 0.504  & 0.292  & 0.210  & \underline{0.264}  & 2.857  & \underline{0.180}  & 0.851  \\
    \midrule
    NsDiff & MAE   & 0.523  & 0.490  & \underline{0.455}  & 0.352  & \underline{0.306}  & 0.412  & \underline{0.985}  & 0.307  & \underline{0.373}  \\
    (2025) & MSE   & 0.594  & 0.514  & \underline{0.488}  & 0.281  & \underline{0.209}  & 0.300  & \underline{2.846}  & 0.242  & \underline{0.637}  \\
    \midrule
    DiffPTS & MAE   & \textbf{0.469} & \textbf{0.476} & \textbf{0.433} & \textbf{0.330} & \textbf{0.263} & \textbf{0.347} & \textbf{0.871} & \textbf{0.234} & \textbf{0.304} \\
    (ours) & MSE   & \textbf{0.507} & \textbf{0.452} & \textbf{0.424} & \textbf{0.242} & \textbf{0.180} & \textbf{0.242} & \textbf{2.636} & \textbf{0.179} & \textbf{0.456} \\
    \bottomrule
    \end{tabular}%
  \label{tab:apdx:exp_rst:full_point}%
\end{table}%

\subsection{Ablation Results with Error Bar}
\label{apdx:subsec:ablation_with_errorbar}

Table~\ref{tab:ablation_errorbar} reports the ablation results with error bars across the nine benchmark datasets. The complete DiffPTS model consistently outperforms all ablated variants on both CRPS and MSE, and removing either NLL component leads to a clear degradation. Removing the NLL of $f_\phi$ (i.e., setting $f_\phi(\mathbf{X}) = \mathbf{0}$) causes the most severe deterioration (e.g., on ETTh1, CRPS increases from 0.374 to 0.553 and MSE from 0.507 to 1.125), since the endpoint then reduces to a zero-mean Gaussian as in standard DDPM. Removing the NLL of $g_\psi$ (i.e., setting $g_\psi(\mathbf{X}) = \mathbf{1}$) leads to moderate but consistent drops (e.g., CRPS increases from 0.374 to 0.416 on ETTh1), showing the benefit of a learned input-dependent variance. The \textit{w/o denoise} variant, which samples directly from the NLL endpoint without the diffusion process, also underperforms the complete model, confirming the necessity of integrating the NLL endpoint with the diffusion mechanism. The small error bars of DiffPTS (mostly within $\pm$0.01 for CRPS and $\pm$0.03 for MSE) demonstrate the stability of our approach.


\begin{table}[htbp]
  \centering
  \footnotesize
  \caption{Ablation results with error bars over seeds $\{1,2,3\}$ on nine datasets. \textbf{Bold face} indicates the best result.}
    \begin{tabular}{ccccccc}
    \toprule
    Variants & Datasets & ETTh1 & ETTh2 & ETTm1 & ETTm2 & - \\
    \midrule
    DiffPTS & CRPS  & \textbf{0.374±0.009} & \textbf{0.355±0.007} & \textbf{0.329±0.005} & \textbf{0.254±0.007} & - \\
          & MSE   & \textbf{0.507±0.033} & \textbf{0.452±0.021} & \textbf{0.424±0.007} & \textbf{0.242±0.031} & - \\
    \midrule
    \multirow{2}{*}{w/o NLL $f_\phi$} & CRPS  & 0.553±0.036 & 0.943±0.013 & 0.540±0.003 & 0.813±0.007 & - \\
          & MSE   & 1.125±0.003 & 3.178±0.053 & 1.101±0.007 & 2.527±0.057 & - \\
    \midrule
   \multirow{2}{*}{w/o NLL $g_\psi$} & CRPS  & 0.416±0.021 & 0.407±0.019 & 0.330±0.008 & 0.279±0.002 & - \\
          & MSE   & 0.631±0.066 & 0.538±0.054 & 0.446±0.021 & 0.294±0.007 & - \\
    \midrule
     \multirow{2}{*}{w/o denoise} & CRPS  & 0.437±0.049 & 0.369±0.024 & 0.387±0.006 & 0.288±0.024 & - \\
          & MSE   & 0.631±0.116 & 0.469±0.085 & 0.498±0.078 & 0.290±0.040 & - \\
    \midrule
    Variants & Datasets & ECL   & EXG   & ILI   & Solar & Traffic \\
    \midrule
    DiffPTS & CRPS  & \textbf{0.218±0.001} & \textbf{0.275±0.012} & \textbf{0.731±0.018} & \textbf{0.173±0.000} & \textbf{0.270±0.001} \\
          & MSE   & \textbf{0.180±0.001} & \textbf{0.242±0.015} & \textbf{2.636±0.073} & \textbf{0.179±0.001} & \textbf{0.456±0.001} \\
    \midrule
    \multirow{2}{*}{w/o NLL $f_\phi$} & CRPS  & 0.544±0.000 & 0.944±0.019 & 1.409±0.019 & 0.494±0.000 & 0.540±0.001 \\
          & MSE   & 1.050±0.000 & 3.231±0.273 & 6.493±0.051 & 0.781±0.000 & 1.431±0.001 \\
    \midrule
    \multirow{2}{*}{w/o NLL $g_\psi$} & CRPS  & 0.389±0.006 & 0.356±0.018 & 0.800±0.018 & 0.324±0.003 & 0.428±0.004 \\
          & MSE   & 0.361±0.006 & 0.312±0.028 & 2.657±0.139 & 0.211±0.009 & 0.653±0.021 \\
    \midrule
    \multirow{2}{*}{w/o denoise} & CRPS  & 0.245±0.006 & 0.346±0.024 & 0.888±0.152 & 0.190±0.008 & 0.277±0.001 \\
          & MSE   & 0.201±0.004 & 0.343±0.036 & 3.271±0.911 & 0.208±0.011 & 0.626±0.007 \\
    \bottomrule
    \end{tabular}%
  \label{tab:ablation_errorbar}%
\end{table}%

\subsection{Full Plug and Play Results}
\label{adpx:subsec:full_plug_play_results}

To further demonstrate the generality and effectiveness of our proposed ELBO formulation, we conduct plug-and-play experiments by integrating the DiffPTS training objective into four existing diffusion-based forecasting models: NsDiff~\cite{ye2025non}, RDIT~\cite{lai2025rdit}, D3U~\cite{li2025diffusion}, and TMDM~\cite{li2024tmdm}. For each base model, we replace its original estimator training strategy with our unified ELBO-driven objective while keeping all other components and hyperparameters unchanged. For D3U and TMDM, we add the same variance backbone we use for DiffPTS for full use of the framework.   As shown in Table~\ref{tab:full_ablation_results}, the ``+DiffPTS'' variants consistently outperform their respective base model. For instance, when applied to D3U on the Traffic dataset, DiffPTS reduces the CRPS from 0.507 to 0.269, a relative improvement of 47.0\%. Similarly, on the ILI dataset, DiffPTS improves D3U's CRPS from 2.097 to 0.762, a reduction of 63.7\%. These concrete results confirm that our principled ELBO reformulation serves as a powerful, architecture-agnostic plug-in module that can significantly enhance the probabilistic forecasting capability of diverse diffusion frameworks.
\begin{table}[htbp]
    \setstretch{0.9}
    \footnotesize
  \centering
  \caption{Plug-and-play results on four methods. \textbf{Bold face} indicates the best result.}
    \begin{tabular}{ccccccccccc}
    \toprule
    Models & Datasets & ETTh1 & ETTh2 & ETTm1 & ETTm2 & ECL   & EXG   & ILI   & Solar & Traffic \\
    \midrule
    TMDM & CRPS  & 0.452  & 0.383  & 0.375  & 0.289  & 0.461  & 0.336  & 0.967  & 0.350  & 0.557  \\
    (2024)    & MSE   & 0.696  & 0.512  & 0.494  & \textbf{0.315 } & 0.257  & 0.334  & 3.636  & 0.250  & 0.679  \\
    \midrule
    \rowcolor{lightgray}  +DiffPTS & CRPS  & \textbf{0.419 } & \textbf{0.380 } & \textbf{0.343 } & \textbf{0.284 } & \textbf{0.358 } & \textbf{0.335 } & \textbf{0.823 } & \textbf{0.306 } & \textbf{0.432 } \\
     \rowcolor{lightgray}  (Ours)    & MSE   & \textbf{0.628 } & \textbf{0.508 } & \textbf{0.476 } & 0.320  & \textbf{0.214 } & \textbf{0.333 } & \textbf{2.976 } & \textbf{0.217 } & \textbf{0.637 } \\
    \midrule
   D3U & CRPS  & 0.447  & 0.385  & 0.412  & 0.282  & 0.439  & 0.354  & 2.097  & 0.488  & 0.507  \\
     (2025)     & MSE   & 0.684  & 0.520  & 0.589  & 0.286  & 0.440  & \textbf{0.269 } & 9.951  & 0.644  & 0.907  \\
    \midrule
    \rowcolor{lightgray} +DiffPTS & CRPS  & \textbf{0.349 } & \textbf{0.331 } & \textbf{0.290 } & \textbf{0.242 } & \textbf{0.222 } & \textbf{0.325 } & \textbf{0.762 } & \textbf{0.181 } & \textbf{0.269 } \\
     \rowcolor{lightgray} (Ours)     & MSE   & \textbf{0.467 } & \textbf{0.369 } & \textbf{0.357 } & \textbf{0.231 } & \textbf{0.187 } & 0.273  & \textbf{2.546 } & \textbf{0.183 } & \textbf{0.457 } \\
    \midrule
    RDIT & CRPS  & 0.410  & 0.380  & 0.336  & 0.258  & 0.362  & 0.392  & 0.845  & 0.225  & 0.495  \\
    (2025)     & MSE   & 0.570  & 0.463  & 0.504  & 0.292  & 0.210  & 0.264  & 2.857  & \textbf{0.180 } & 0.851  \\
    \midrule
    \rowcolor{lightgray} +DiffPTS & CRPS  & \textbf{0.338 } & \textbf{0.363 } & \textbf{0.320 } & \textbf{0.245 } & \textbf{0.231 } & \textbf{0.292 } & \textbf{0.807 } & \textbf{0.222 } & \textbf{0.289 } \\
    \rowcolor{lightgray} (Ours)     & MSE   & \textbf{0.458 } & \textbf{0.441 } & \textbf{0.375 } & \textbf{0.216 } & \textbf{0.186 } & \textbf{0.254 } & \textbf{2.714 } & 0.200  & \textbf{0.591 } \\
    \midrule
    NsDiff & CRPS  & 0.392  & 0.358  & 0.346  & 0.256  & 0.290  & 0.324  & 0.806  & 0.300  & 0.378  \\
     (2025)     & MSE   & 0.594  & 0.514  & 0.488  & 0.281  & 0.209  & 0.300  & 2.846  & 0.242  & 0.637  \\
    \midrule
    \rowcolor{lightgray} +DiffPTS & CRPS  & \textbf{0.364 } & \textbf{0.355 } & \textbf{0.339 } & \textbf{0.254 } & \textbf{0.218 } & \textbf{0.275 } & \textbf{0.761 } & \textbf{0.199 } & \textbf{0.275 } \\
    \rowcolor{lightgray} (Ours)      & MSE   & \textbf{0.576 } & \textbf{0.495 } & \textbf{0.472 } & \textbf{0.278 } & \textbf{0.199 } & \textbf{0.280 } & \textbf{2.739 } & \textbf{0.232 } & \textbf{0.587 } \\
    \bottomrule
    \end{tabular}%
  \label{tab:full_ablation_results}%
\end{table}%

\subsection{Efficiency Analysis}
\label{apdx:efficiency_analysis}
To analyze the computational efficiency of the proposed framework DiffPTS, we compare the training time, memory consumption, and downstream performance, with results reported in Table~\ref{tab:computational_efficiency}. The computational efficiency of DiffPTS is striking: with only 8MB additional memory, our framework achieves an average 55.3\% reduction in training time and 15.6\% improvement in CRPS across TMDM, D3U, and RDIT. These gains stem from our key insight that \textit{pretraining can be eliminated through proper joint optimization}. The proposed ELBO formulation enables end-to-end training that is not only more efficient but also yields superior predictive performance, challenging the conventional two-stage training paradigm in diffusion-based time series forecasting.
\begin{table}[htbp]
  \centering
  \footnotesize
  \caption{Efficiency analysis results of DiffPTS on ETTh1 dataset.}
    \begin{tabular}{cccccc}
    \toprule
          & Memory & Average Epoch Time & Training Strategy & CRPS  & {MSE} \\
    \midrule
    TMDM  & 906MB & 162.74ms & pretraining+DDPM & 0.452  & {0.696} \\
   \rowcolor{lightgray} +DiffPTS & 914MB & 76.61ms & joint optimization & \textbf{0.419} & \textbf{0.628} \\
    \midrule
    D3U   & 675MB & 130.40ms & pretraining+DDPM & 0.447  & {0.684} \\
    \rowcolor{lightgray}+DiffPTS & 683MB & 54.78ms & joint optimization & \textbf{0.349} & \textbf{0.467} \\
    \midrule
    RDIT  & 490MB & 105.75ms & pretraining+DDPM & 0.410  & {0.570} \\
  \rowcolor{lightgray}  +DiffPTS & 498MB & 47.66ms & joint optimization & \textbf{0.338} & \textbf{0.458} \\
    \midrule
    NsDiff & 754MB & 234.15ms & pretraining+DDPM & 0.392  & 0.594 \\
   \rowcolor{lightgray} +DiffPTS & 754MB & 106.25ms & joint optimization & \textbf{0.364} & \textbf{0.576} \\
    \bottomrule
    \end{tabular}%
  \label{tab:computational_efficiency}%
\end{table}%

\section{Further Analyses and Discussions}
\label{apdx:sec:further}

This section provides further analyses and discussions of the modeling assumption, the training objective, the estimators, and the sampling procedure. Unless otherwise stated, all experiments follow the setting of Section~\ref{subsec:exp_setup} with seeds $\{1,2,3\}$.

\subsection{On the LSNM Assumption and the Diagonal Covariance}
\label{apdx:subsec:lsnm}

Following prior work~\cite{han2022card, ye2025non}, the LSNM is often written directly on the target as $\mathbf{Y} = f_\phi(\mathbf{X}) + \sqrt{g_\psi(\mathbf{X})} \odot \boldsymbol{\eta}$. Strictly speaking, however, in the diffusion models considered here the LSNM specifies the endpoint of the diffusion process rather than the final prediction, i.e., $p(\mathbf{Y}_T \mid \mathbf{X}) = \mathcal{N}(f_\phi(\mathbf{X}), g_\psi(\mathbf{X}))$ as stated in Equation~\eqref{eq:LSNM}. The forward process in Equation~\eqref{eq:forward_Yt_Nsdiff} is constructed such that $q(\mathbf{Y}_T \mid \mathbf{Y}_0, \mathbf{X})$ approaches this endpoint as $\bar{\alpha}_T$ decreases, which is why the assumption is commonly stated on $\mathbf{Y}$ itself. The diagonal Gaussian with covariance $\mathrm{diag}(g_\psi(\mathbf{X}))$ therefore only initializes the generative process. The final predictive distribution is the full reverse process $p_\theta(\mathbf{Y}_0 \mid \mathbf{X})$, whose transitions are conditioned on the entire history $\mathbf{X}$ and on all coordinates of $\mathbf{Y}_t$ and iteratively refine the sample. It is therefore not restricted to a diagonal Gaussian, and dependencies across time steps and variates can be captured through the reverse chain. A diagonal endpoint is also the standard setting of all compared methods, namely DDPM, NsDiff, TMDM, RDIT, and D3U. Explicitly modeling the full conditional covariance within the ELBO is an interesting direction for future work.

\subsection{Joint versus Two-Phase Training}
\label{apdx:subsec:joint_vs_two_phase}

\textbf{Architectural differences.} Table~\ref{tab:apdx:arch_comparison} summarizes the components of DiffPTS in Table~\ref{tab:main_results} and of the hosts and their +DiffPTS variants in Table~\ref{tab:plugin_play_results}. All +DiffPTS variants are trained end-to-end with the same joint objective as DiffPTS (Equation~\eqref{eq:denoise_proposition_diffpts}). Within each host, the architecture is held fixed and only the training procedure changes, namely the original two-phase or auxiliary training is replaced by joint ELBO training. DiffPTS in Table~\ref{tab:main_results} deliberately inherits its $f_\phi$ and $g_\psi$ architectures from NsDiff (a Non-stationary Transformer~\cite{liu2022non} and a three-layer MLP), so that the comparison against the two-phase baselines isolates the training objective rather than the architecture. D3U instead contributes a stronger patch-based mean backbone~\cite{nie2022time}, which is known to be particularly strong on the ETT datasets, and D3U+DiffPTS benefits from it. The gap between DiffPTS and D3U+DiffPTS on the ETT datasets is therefore architectural. The estimators themselves are not our contribution. Our contribution is how they should be trained, namely the ELBO itself prescribes a Gaussian NLL objective for whatever $f_\phi$ and $g_\psi$ are chosen (Proposition~\ref{prop:loss_function_DiffPTS}), which is why the same objective improves every host.

\begin{table}[htbp]
  \centering
  \footnotesize
  \caption{Components of DiffPTS (Table~\ref{tab:main_results}) and of the plug-and-play hosts with their +DiffPTS variants (Table~\ref{tab:plugin_play_results}). Joint ELBO denotes end-to-end training with Equation~\eqref{eq:denoise_proposition_diffpts}.}
    \begin{tabular}{lcccc}
    \toprule
    Model & Training & Mean estimator $f_\phi$ & Variance estimator $g_\psi$ & Noise schedule \\
    \midrule
    DiffPTS (Table~\ref{tab:main_results}) & Joint ELBO & NS-Transformer & MLP (NLL) & Standard linear \\
    \midrule
    NsDiff & Two-phase & NS-Transformer & MLP (sliding-window target) & Uncertainty-aware \\
    NsDiff+DiffPTS & Joint ELBO & NS-Transformer & MLP (NLL) & Standard linear \\
    \midrule
    RDIT & Two-phase & Mamba~\cite{gu2023mamba} & Fixed training residual & Standard linear \\
    RDIT+DiffPTS & Joint ELBO & Mamba~\cite{gu2023mamba} & MLP (NLL) & Standard linear \\
    \midrule
    D3U & Two-phase & Patch-based & None (implicit $g_\psi \equiv 1$) & Standard linear \\
    D3U+DiffPTS & Joint ELBO & Patch-based & Added MLP (NLL) & Standard linear \\
    \bottomrule
    \end{tabular}%
  \label{tab:apdx:arch_comparison}%
\end{table}%

\textbf{Two-phase training under an identical architecture.} To directly test the training objective, we compare DiffPTS with a two-phase variant that shares its architecture, noise schedule, and seeds. In this variant, $f_\phi$ is pretrained with MSE and $g_\psi$ with the future sliding-window variance target of NsDiff, both are then frozen, and only the denoiser is trained. As reported in Table~\ref{tab:apdx:two_phase}, joint ELBO training improves CRPS/MSE from 0.404/0.647 to 0.374/0.507 on ETTh1 (7.4\%/21.6\%) and from 0.331/0.453 to 0.329/0.424 on ETTm1, while removing the pretraining stage entirely. This confirms that the gains come from the training objective rather than from the architecture.

\begin{table}[htbp]
  \centering
  \footnotesize
  \caption{Joint versus two-phase training under an identical architecture, noise schedule, and seeds (CRPS / MSE).}
    \begin{tabular}{lcc}
    \toprule
    Training & ETTh1 & ETTm1 \\
    \midrule
    Two-phase (pretrain, then freeze $f_\phi$ and $g_\psi$) & 0.404 / 0.647 & 0.331 / 0.453 \\
    Joint ELBO (DiffPTS) & \textbf{0.374} / \textbf{0.507} & \textbf{0.329} / \textbf{0.424} \\
    \bottomrule
    \end{tabular}%
  \label{tab:apdx:two_phase}%
\end{table}%

\textbf{Differences between the estimators of DiffPTS and D3U.} As summarized in Table~\ref{tab:method_comparison}, D3U introduces no variance estimator. It diffuses the residual $\mathbf{Y} - f_\phi(\mathbf{X})$ toward $\mathcal{N}(\mathbf{0}, \mathbf{I})$, which amounts to an implicit fixed unit variance in the residual space, whereas DiffPTS learns an input-dependent $g_\psi(\mathbf{X})$ through the ELBO. Both methods estimate the conditional mean $\mathbb{E}[\mathbf{Y} \mid \mathbf{X}]$ with $f_\phi$, and the difference lies in the training signal. D3U pretrains $f_\phi$ with a plain MSE and then freezes it, while DiffPTS trains $f_\phi$ jointly through the NLL term, whose squared error is weighted by $1/g_\psi(\mathbf{X})$, so that $f_\phi$ allocates its capacity according to the predicted uncertainty. In addition, $f_\phi$ receives gradients from the denoising term, since $\mathbf{Y}_t$ in Equation~\eqref{eq:forward_Yt_Nsdiff} depends on it. These differences are visible in Figure~\ref{fig:sample_showcase}. The intervals of D3U and TMDM have a nearly constant width over time as a consequence of the missing variance estimator (Figure~\ref{fig:sample_showcase}(e-f)), while the intervals of DiffPTS track the empirical intervals (Figure~\ref{fig:sample_showcase}(a-b)), and Figure~\ref{fig:sample_showcase}(g) compares the estimated densities directly. Quantitatively, the calibration of the predictive distributions is reported as QICE in Table~\ref{tab:apdx:exp_rst:full_probabilistic}, where DiffPTS is the best or second best on every dataset, while D3U suffers from its fixed-variance assumption.

\subsection{Stationary versus Non-Stationary Datasets}
\label{apdx:subsec:stationarity}

\textbf{Intuition.} When a dataset is near-stationary, the conditional mean and variance are stable functions of the input, so both estimators are easy to fit, and in the limit of a constant conditional variance, $g_\psi$ degenerates toward the fixed-variance special case $g_\psi \equiv 1$ of D3U and TMDM (Remark~\ref{remark:model_equivalence}). Under non-stationarity, the location and scale that $f_\phi$ and $g_\psi$ must track drift over time. Externally supervised targets, such as the sliding-window variance of NsDiff or the training residuals of RDIT, then become mis-specified, whereas the ELBO-prescribed NLL requires no external target and adapts the estimators to each input. A particularly favorable regime is periodic heteroscedastic data such as Solar and Traffic, where the variance changes strongly within a day yet is highly predictable from the input, which is where a learned $g_\psi(\mathbf{X})$ helps most.

\textbf{Quantitative analysis.} We compare each host with its own +DiffPTS variant using the full plug-and-play results in Table~\ref{tab:full_ablation_results}, so that every method is measured against its own baseline. To classify the datasets, we standardize each dataset with the statistics of its training split, slide windows of length 168, and take the largest deviation across variates of the window mean from 0 (mean shift) and of the window variance from 1 (variance shift), since a single strongly drifting variate already requires non-stationary modeling. Splitting at the median yields a non-stationary group (Traffic, EXG, ILI, ECL, and ETTm2) and a stationary group (ETTh2, ETTm1, ETTh1, and Solar). Table~\ref{tab:apdx:stationarity} reports the relative CRPS improvement of +DiffPTS over each host. The gains of the ELBO objective are positive everywhere and consistently larger on the non-stationary group for all four hosts, in line with the intuition that externally supervised estimator targets are most mis-specified under distribution shift.

\begin{table}[htbp]
  \centering
  \footnotesize
  \caption{Relative CRPS improvement of +DiffPTS over each host, grouped by non-stationarity. The mean and variance shifts are the largest deviations of the sliding-window mean from 0 and of the sliding-window variance from 1 after standardization, where larger values indicate stronger non-stationarity.}
    \begin{tabular}{llccccc}
    \toprule
    Group & Dataset & Mean / variance shift & TMDM & D3U & RDIT & NsDiff \\
    \midrule
    \multirow{6}{*}{Non-stationary} & Traffic & 1.59 / 35.4 & 22.4\% & 46.9\% & 41.6\% & 27.2\% \\
     & EXG & 1.45 / 0.98 & 0.3\% & 8.2\% & 25.5\% & 15.1\% \\
     & ILI & 1.02 / 1.18 & 14.9\% & 63.7\% & 4.5\% & 5.6\% \\
     & ECL & 0.89 / 1.26 & 22.3\% & 49.4\% & 36.2\% & 24.8\% \\
     & ETTm2 & 1.03 / 1.00 & 1.7\% & 14.2\% & 5.0\% & 0.8\% \\
     & Average & -- & \textbf{12.3\%} & \textbf{36.5\%} & \textbf{22.6\%} & \textbf{14.7\%} \\
    \midrule
    \multirow{5}{*}{Stationary} & ETTh2 & 1.02 / 1.01 & 0.8\% & 14.0\% & 4.5\% & 0.8\% \\
     & ETTm1 & 0.89 / 0.95 & 8.5\% & 29.6\% & 4.8\% & 2.0\% \\
     & ETTh1 & 0.86 / 0.90 & 7.3\% & 21.9\% & 17.6\% & 7.1\% \\
     & Solar & 0.25 / 0.37 & 12.6\% & 62.9\% & 1.3\% & 33.7\% \\
     & Average & -- & 7.3\% & 32.1\% & 7.1\% & 10.9\% \\
    \bottomrule
    \end{tabular}%
  \label{tab:apdx:stationarity}%
\end{table}%

\subsection{Effect of the ELBO Weighting Coefficients}
\label{apdx:subsec:weighting}

The coefficients $\gamma_t$ and $\bar{\alpha}_T$ in Proposition~\ref{proposition:complete_elbo} are not tunable hyperparameters. Both are closed-form constants fully determined by the noise schedule, with $\gamma_t = \beta_t/(2\alpha_t(1-\bar{\alpha}_{t-1}))$ for $t>1$, $\gamma_1 = 1/(2\alpha_1)$, and $\bar{\alpha}_T = \prod_{t=1}^{T}\alpha_t$. Under our schedule ($T=20$, linear $\beta_t$ from $10^{-4}$ to $0.02$), $\gamma_t$ ranges over $[0.06, 5.74]$ and $\bar{\alpha}_T \approx 0.82$, and no tuning is performed anywhere in the paper. These constants weight the per-step denoising terms and the quadratic endpoint term of the exact ELBO. Following the standard practice of DDPM~\cite{ho2020denoising}, our implementation uses the simplified unweighted objective of Proposition~\ref{prop:loss_function_DiffPTS}, which drops these constants. We therefore refer to Equation~\eqref{eq:denoise_proposition_diffpts} as the ELBO-derived objective with the standard simplified weighting, rather than as the exact ELBO. To quantify the effect of the weights, we train with the exact-weighted objective, which places $\gamma_t$ on the denoising terms and $\bar{\alpha}_T/2$ on the quadratic NLL term, and compare it with the simplified objective in Table~\ref{tab:apdx:weighting}. The simplified objective is equal or better on every dataset, mirroring the finding of Ho et al.~\cite{ho2020denoising}. We also note that $\bar{\alpha}_T \approx 0.82$ gives the endpoint NLL term a large weight in the exact ELBO under our setting, consistent with the ablations in Table~\ref{tab:ablation_main}, where removing the NLL components substantially degrades performance.

\begin{table}[htbp]
  \centering
  \footnotesize
  \caption{Exact-weighted versus simplified ELBO objective (CRPS / MSE).}
    \begin{tabular}{lccc}
    \toprule
    Objective & ETTh1 & ETTm1 & EXG \\
    \midrule
    Exact-weighted ELBO ($\gamma_t$, $\bar{\alpha}_T/2$) & 0.406 / 0.629 & 0.330 / 0.455 & 0.281 / 0.275 \\
    Simplified weighting (DiffPTS) & \textbf{0.374} / \textbf{0.507} & \textbf{0.329} / \textbf{0.424} & \textbf{0.275} / \textbf{0.242} \\
    \bottomrule
    \end{tabular}%
  \label{tab:apdx:weighting}%
\end{table}%

\subsection{Numerical Stability of the Gaussian NLL}
\label{apdx:subsec:stability}

A very small $g_\psi(\mathbf{X})$ might be expected to cause instability in the NLL term $(\mathbf{Y}_0 - f_\phi(\mathbf{X}))^2/g_\psi(\mathbf{X}) + \log g_\psi(\mathbf{X})$. This term is, however, self-balancing. As $g_\psi$ approaches zero, the quadratic term diverges, which prevents variance collapse, while the logarithmic term prevents the variance from blowing up. The implementation additionally uses a softplus output (Appendix~\ref{apdx:subsec:detail_implementation}) and a small variance floor of $10^{-8}$. Across all nine datasets, each with three seeds, we observed no NaN or divergence. Table~\ref{tab:apdx:stability} reports test-set statistics of the learned variance on ETTh1 and ETTm1. The learned $g_\psi$ stays well bounded away from zero, whereas the hand-crafted alternative is the brittle one, in the sense that the sliding-window variance target of NsDiff degenerates toward zero on locally flat segments and its Gaussian NLL explodes there. This further supports learning $g_\psi$ through the ELBO.

\begin{table}[htbp]
  \centering
  \footnotesize
  \caption{Test-set statistics of the learned variance $g_\psi$ and of the sliding-window variance target of NsDiff, computed on the ground-truth future.}
    \begin{tabular}{lcc}
    \toprule
    Statistic & ETTh1 & ETTm1 \\
    \midrule
    Learned $g_\psi$, minimum & $1.6\times10^{-2}$ & $5.4\times10^{-3}$ \\
    Learned $g_\psi$, median & 0.11 & 0.10 \\
    Learned $g_\psi$, maximum & 4.0 & 2.9 \\
    \midrule
    Sliding-window target, minimum & $2.2\times10^{-5}$ & $6.0\times10^{-7}$ \\
    Gaussian NLL with the sliding-window target & 0.32 & $1.1\times10^{4}$ \\
    \bottomrule
    \end{tabular}%
  \label{tab:apdx:stability}%
\end{table}%

\subsection{Choice of the Number of Diffusion Steps}
\label{apdx:subsec:T}

Our choice of $T=20$ follows the convention of prior diffusion forecasters~\cite{rasul2021autoregressive, ye2025non}. In particular, TimeGrad~\cite{rasul2021autoregressive} reports that its performance already saturates around $T=20$. We observe a similar saturation and further find that an overly large $T$ can even hurt performance, while a very small $T$ is only slightly worse (Table~\ref{tab:apdx:T}). $T=20$ is the best on all datasets. We explain this behavior by answering two questions below.

\begin{table}[htbp]
  \centering
  \footnotesize
  \caption{CRPS of DiffPTS under different numbers of diffusion steps $T$. All entries come from a separate set of runs under an identical configuration, so the $T=20$ column may differ from Table~\ref{tab:main_results}.}
    \begin{tabular}{lcccc}
    \toprule
    Dataset & $T=1$ & $T=20$ & $T=100$ & $T=1000$ \\
    \midrule
    ETTh1 & 0.387 & \textbf{0.377} & 0.432 & 0.552 \\
    ECL   & 0.232 & \textbf{0.218} & 0.259 & 0.498 \\
    ETTm2 & 0.284 & \textbf{0.274} & 0.353 & 0.554 \\
    EXG   & 0.307 & \textbf{0.275} & 0.320 & 0.639 \\
    \bottomrule
    \end{tabular}%
  \label{tab:apdx:T}%
\end{table}%

\textbf{Why $T$ can be much smaller for time series than for images.} Each reverse transition is a small Gaussian-approximated correction. Images are very high-dimensional (e.g., $256\times256\times3 \approx 2\times10^5$ dimensions) and their distributions are sharply multimodal, so many small steps are needed to fit them, typically around $T=1000$~\cite{ho2020denoising}. Forecasting targets are far lower dimensional (e.g., $192\times7 \approx 1.3\times10^3$ dimensions on the ETT datasets, and TimeGrad even diffuses a single time step at a time). More importantly, the target is a conditional distribution in which the history $\mathbf{X}$ and the target $\mathbf{Y}$ are often highly similar, so the history already pins down most of the structure and leaves a close-to-unimodal conditional distribution to model. This also explains why TimeGrad, whose endpoint is a fixed $\mathcal{N}(\mathbf{0}, \mathbf{I})$, saturates at around $T=20$ to $100$. The learned endpoint $\mathcal{N}(f_\phi(\mathbf{X}), g_\psi(\mathbf{X}))$ of DiffPTS shortens the path further by absorbing the location and scale, leaving only mild non-Gaussian structure to the reverse chain. This is visible in Figure~\ref{fig:overview} and Figure~\ref{fig:sample_showcase}(g), where the learned distribution differs from the Gaussian endpoint yet only mildly reshapes it and remains unimodal. Empirically, reducing $T$ from 20 to 1 on ETTh1 loses little (0.377 to 0.387), whereas removing the denoising entirely clearly degrades performance (0.374 to 0.437 in Table~\ref{tab:ablation_main}), confirming that only a small number of corrective steps is needed.

\textbf{Why $T$ should not be too large.} We conjecture that the main cause lies in the balance between model complexity and data scarcity. A larger $T$ multiplies the model complexity, since the time-conditioned denoiser must fit $T$ distinct denoising sub-tasks and thus needs more data. Time series data are scarce (a single trajectory per dataset, e.g., fewer than 1000 steps on ILI), and under non-stationarity different regimes follow different distributions, which further reduces the effective sample size per regime. A large $T$ therefore increases the estimation burden without adding useful capacity, consistent with the degradation in Table~\ref{tab:apdx:T}.

In summary, the low-dimensional prediction target, the non-stationarity, and the data scarcity of time series forecasting jointly call for a compromise between a small and a large $T$. The optimal $T$ is not known a priori, which is an interesting direction for future study, yet prior work~\cite{rasul2021autoregressive, ye2025non} and our results suggest that a value around 20 is a good choice.

\subsection{Fast Sampling with DPM-Solver}
\label{apdx:subsec:dpm}

Although DiffPTS simplifies training, inference still iterates the reverse diffusion. DiffPTS nevertheless composes directly with fast training-free samplers~\cite{song2021denoising, liu2022pseudo, lu2022dpm, lu2025dpmpp}. Under the change of variables $\mathbf{Z} = (\mathbf{Y} - f_\phi(\mathbf{X}))/\sqrt{g_\psi(\mathbf{X})}$ of Appendix~\ref{sec:proof:model_equivalence}, the forward process in Equation~\eqref{eq:forward_Yt_Nsdiff} becomes exactly a standard variance-preserving diffusion,
\begin{equation}
\mathbf{Z}_t = \sqrt{\bar{\alpha}_t}\, \mathbf{Z}_0 + \sqrt{1-\bar{\alpha}_t}\, \boldsymbol{\epsilon}, \qquad \mathbf{Z}_T \sim \mathcal{N}(\mathbf{0}, \mathbf{I}),
\end{equation}
driven by the same noise $\boldsymbol{\epsilon}$ that $\boldsymbol{\epsilon}_\theta$ is trained to predict. Standard ODE solvers can therefore be applied in $\mathbf{Z}$-space without retraining. The learned endpoint $\mathcal{N}(f_\phi(\mathbf{X}), g_\psi(\mathbf{X}))$ corresponds exactly to the standard initial state $\mathcal{N}(\mathbf{0}, \mathbf{I})$ in $\mathbf{Z}$-space, and samples are mapped back via $\mathbf{Y} = f_\phi(\mathbf{X}) + \sqrt{g_\psi(\mathbf{X})} \odot \mathbf{Z}$. We evaluate the second-order multistep DPM-Solver~\cite{lu2022dpm} with 100 samples and seeds $\{1,2,3\}$, where all rows of Table~\ref{tab:apdx:dpm} use the same checkpoints and evaluation protocol. With only 5 steps, DPM-Solver stays within about 3\% CRPS of ancestral sampling at roughly $2\times$ speedup, so DiffPTS remains effective under fast sampling.

\begin{table}[htbp]
  \centering
  \footnotesize
  \caption{Ancestral sampling versus DPM-Solver for DiffPTS (CRPS / seconds per batch). All rows use the same checkpoints and a separate evaluation script, so values are comparable within this table.}
    \begin{tabular}{lcc}
    \toprule
    Sampler & ETTm1 & ETTh1 \\
    \midrule
    Ancestral, $T=20$ (Algorithm~\ref{alg:diffpts_sampling}) & 0.333 / 1.31s & 0.422 / 0.68s \\
    DPM-Solver, 10 steps & 0.340 / 0.85s & 0.436 / 0.64s \\
    DPM-Solver, 5 steps & 0.340 / 0.61s & 0.436 / 0.32s \\
    \bottomrule
    \end{tabular}%
  \label{tab:apdx:dpm}%
\end{table}%

\subsection{$x_0$-Prediction versus $\epsilon$-Prediction}
\label{apdx:subsec:x0}

The two parameterizations are equivalent up to the affine reparameterization of Equation~\eqref{eq:forward_Yt_Nsdiff} with a correspondingly reweighted loss. Inverting Equation~\eqref{eq:forward_Yt_Nsdiff} gives
\begin{equation}
\hat{\mathbf{Y}}_0 = \frac{1}{\sqrt{\bar{\alpha}_t}}\Big(\mathbf{Y}_t - \big(1-\sqrt{\bar{\alpha}_t}\big) f_\phi(\mathbf{X}) - \sqrt{(1-\bar{\alpha}_t)\, g_\psi(\mathbf{X})} \odot \boldsymbol{\epsilon}_\theta\Big),
\end{equation}
so that the two objectives satisfy
\begin{equation}
\big\|\mathbf{Y}_0 - \hat{\mathbf{Y}}_0\big\|^2 = \frac{1-\bar{\alpha}_t}{\bar{\alpha}_t} \big\|\sqrt{g_\psi(\mathbf{X})} \odot (\boldsymbol{\epsilon}_0 - \boldsymbol{\epsilon}_\theta)\big\|^2,
\end{equation}
i.e., the $x_0$-prediction loss is an $\epsilon$-prediction loss reweighted by $(1-\bar{\alpha}_t)/\bar{\alpha}_t$ and $g_\psi(\mathbf{X})$. We train an $x_0$-prediction variant, in which the network predicts $\mathbf{Y}_0$ and is plugged into the same reverse posterior, on ETTh1 and ETTm2 (Table~\ref{tab:apdx:x0}). Training remains stable under $x_0$-prediction, but $\epsilon$-prediction is clearly preferable at our small $T=20$, consistent with common observations for small numbers of diffusion steps. We therefore adopt $\epsilon$-prediction.

\begin{table}[htbp]
  \centering
  \footnotesize
  \caption{$x_0$-prediction versus $\epsilon$-prediction for DiffPTS (CRPS / MSE).}
    \begin{tabular}{lcc}
    \toprule
    Parameterization & ETTh1 & ETTm2 \\
    \midrule
    $x_0$-prediction & 0.513 / 0.778 & 0.314 / 0.341 \\
    $\epsilon$-prediction (DiffPTS) & \textbf{0.374} / \textbf{0.507} & \textbf{0.254} / \textbf{0.242} \\
    \bottomrule
    \end{tabular}%
  \label{tab:apdx:x0}%
\end{table}%

\subsection{Limitations and Failure Cases}
\label{apdx:subsec:limitations}

Beyond the limitations discussed in Section~\ref{sec:conclusion}, we summarize the cases where DiffPTS helps less. First, the gains of the ELBO objective are marginal when a host's estimators are already well specified, e.g., 0.8\% CRPS for NsDiff on ETTh2 and ETTm2 (Table~\ref{tab:apdx:stationarity}), and in the limit of a constant conditional variance, $g_\psi$ degenerates toward the fixed-variance case $g_\psi \equiv 1$ of D3U and TMDM. Second, DiffPTS is not uniformly the best on every metric. NsDiff attains a better QICE on ETTh2 and ILI (Table~\ref{tab:apdx:exp_rst:full_probabilistic}), which we attribute to its uncertainty-aware schedule when the sliding-window variance is a good approximation, and a few plug-and-play variants have a slightly higher MSE than their hosts on individual datasets, namely TMDM on ETTm2, D3U on EXG, and RDIT on Solar (Table~\ref{tab:full_ablation_results}). Third, the LSNM endpoint uses a diagonal covariance (Appendix~\ref{apdx:subsec:lsnm}), we follow the simplified weighting of DDPM (Appendix~\ref{apdx:subsec:weighting}), and the number of diffusion steps is chosen empirically (Appendix~\ref{apdx:subsec:T}).

\clearpage
\newpage
\section*{NeurIPS Paper Checklist}

\begin{enumerate}

\item {\bf Claims}
    \item[] Question: Do the main claims made in the abstract and introduction accurately reflect the paper's contributions and scope?
    \item[] Answer: \answerYes{} 
    \item[] Justification: Our claim is verified by Table~\ref{tab:method_comparison} and Proposition~\ref{proposition:complete_elbo}. 

\item {\bf Limitations}
    \item[] Question: Does the paper discuss the limitations of the work performed by the authors?
    \item[] Answer: \answerYes{} 
    \item[] Justification: We discussed the limitation in Section~\ref{sec:conclusion}.

\item {\bf Theory assumptions and proofs}
    \item[] Question: For each theoretical result, does the paper provide the full set of assumptions and a complete (and correct) proof?
    \item[] Answer: \answerYes{} 
    \item[] Justification: The full assumption and proof is provided in Section~\ref{apdx:sec:deriviation} 

    \item {\bf Experimental result reproducibility}
    \item[] Question: Does the paper fully disclose all the information needed to reproduce the main experimental results of the paper to the extent that it affects the main claims and/or conclusions of the paper (regardless of whether the code and data are provided or not)?
    \item[] Answer: \answerYes{} 
    \item[] Justification: Experiment setup and implementation are discussed in Section~\ref{sec:Experiments}. 

\item {\bf Open access to data and code}
    \item[] Question: Does the paper provide open access to the data and code, with sufficient instructions to faithfully reproduce the main experimental results, as described in supplemental material?
    \item[] Answer: \answerYes{} 
    \item[] Justification: The code is publicly available at \url{https://github.com/wwy155/DiffPTS}, and all datasets are public benchmarks (Appendix~\ref{apdx:subsec:dataset_details}).

\item {\bf Experimental setting/details}
    \item[] Question: Does the paper specify all the training and test details (e.g., data splits, hyperparameters, how they were chosen, type of optimizer) necessary to understand the results?
    \item[] Answer: \answerYes{} 
    \item[] Justification: All the training and test details are discussed in Section~\ref{subsec:exp_setup} and Appendix~\ref{apdx:subsec:detail_implementation}. 

\item {\bf Experiment statistical significance}
    \item[] Question: Does the paper report error bars suitably and correctly defined or other appropriate information about the statistical significance of the experiments?
    \item[] Answer: \answerYes{} 
    \item[] Justification: The results with error bars are provided in Appendix~\ref{apdx:subsec:ablation_with_errorbar}

\item {\bf Experiments compute resources}
    \item[] Question: For each experiment, does the paper provide sufficient information on the computer resources (type of compute workers, memory, time of execution) needed to reproduce the experiments?
    \item[] Answer: \answerYes{}  
    \item[] Justification: We discuss the hardware and other details in Section~\ref{subsec:exp_setup}, and an efficiency analysis is provided in Appendix~\ref{apdx:efficiency_analysis} to justify the required computational cost.
    
\item {\bf Code of ethics}
    \item[] Question: Does the research conducted in the paper conform, in every respect, with the NeurIPS Code of Ethics \url{https://neurips.cc/public/EthicsGuidelines}?
    \item[] Answer:\answerYes{} 
    \item[] Justification: The research conforms with the NeurIPS Code of Ethics in every respect.

\item {\bf Broader impacts}
    \item[] Question: Does the paper discuss both potential positive societal impacts and negative societal impacts of the work performed?
    \item[] Answer: \answerNA{} 
    \item[] Justification: The paper focuses solely on technical contributions without discussing potential societal impacts, either positive or negative.
    
\item {\bf Safeguards}
    \item[] Question: Does the paper describe safeguards that have been put in place for responsible release of data or models that have a high risk for misuse (e.g., pre-trained language models, image generators, or scraped datasets)?
    \item[] Answer: \answerNA{} 
    \item[] Justification: This work does not involve the release of high-risk data or models.

\item {\bf Licenses for existing assets}
    \item[] Question: Are the creators or original owners of assets (e.g., code, data, models), used in the paper, properly credited and are the license and terms of use explicitly mentioned and properly respected?
    \item[] Answer: \answerYes{} 
    \item[] Justification: The creators of all datasets used in this work are properly credited with citations provided in Appendix~\ref{apdx:subsec:dataset_details}.

\item {\bf New assets}
    \item[] Question: Are new assets introduced in the paper well documented and is the documentation provided alongside the assets?
    \item[] Answer: \answerYes{} 
    \item[] Justification: The model code repository in the abstract is the only new asset introduced in this work. We provide the code and a documentation in the code repository.

\item {\bf Crowdsourcing and research with human subjects}
    \item[] Question: For crowdsourcing experiments and research with human subjects, does the paper include the full text of instructions given to participants and screenshots, if applicable, as well as details about compensation (if any)? 
    \item[] Answer: \answerNA{}
    \item[] Justification: This paper does not involve crowdsourcing experiments or research with human subjects.

\item {\bf Institutional review board (IRB) approvals or equivalent for research with human subjects}
    \item[] Question: Does the paper describe potential risks incurred by study participants, whether such risks were disclosed to the subjects, and whether Institutional Review Board (IRB) approvals (or an equivalent approval/review based on the requirements of your country or institution) were obtained?
    \item[] Answer:\answerNA{} 
    \item[] Justification: This paper does not involve research with human subjects, therefore IRB approval is not applicable.

\item {\bf Declaration of LLM usage}
    \item[] Question: Does the paper describe the usage of LLMs if it is an important, original, or non-standard component of the core methods in this research? Note that if the LLM is used only for writing, editing, or formatting purposes and does \emph{not} impact the core methodology, scientific rigor, or originality of the research, declaration is not required.
    \item[] Answer: \answerNA{} 
    \item[] Justification: The core method development in this research does not involve LLMs as any important, original, or non-standard components.

\end{enumerate}

\end{document}